\pdfoutput=1
\documentclass[10pt, a4paper]{article}
\usepackage[dvipsnames]{xcolor}

\usepackage[preprint]{acl}

\usepackage{times}
\usepackage{latexsym}
\usepackage{tikz}
\usetikzlibrary{positioning}

\usepackage{microtype}

\usepackage{inconsolata}

\usepackage{graphicx}

\usepackage{booktabs}
\usepackage{multirow}

\title{On the Credibility of Evaluating LLMs using Survey Questions}

\author{Jindřich Libovický \\
Charles University, Faculty of Mathematics and Physics, \\
        Institute of Formal and Applied Linguistics \\
         V~Holešovičkách 747/2, 180 00 Praha, Czechia \\
         \texttt{libovicky@ufal.mff.cuni.cz} \\}

\usepackage{xcolor}
\definecolor{pltblue}{RGB}{31, 77, 180}
\definecolor{pltgreen}{RGB}{214, 39, 40}

\usepackage{etoolbox}
\usepackage{pgf}
\usepackage{colortbl}

\newcommand{\gradientcell}[6]{%
    \def\value{#1}%
    \def\minvalue{#2}%
    \def\maxvalue{#3}%
    \def\mincolor{#4}%
    \def\maxcolor{#5}%
    \def\transparency{#6}%
    \ifdimcomp{\value pt}{>}{\maxvalue pt}{\cellcolor{#5!100.0!#4!#6}\value}{%
    \ifdimcomp{\value pt}{<}{\minvalue pt}{\cellcolor{#5!0.0!#4!#6}\value}{%
         \pgfmathparse{int(round(100*(#1/(\maxvalue-\minvalue))-(\minvalue *(100/(\maxvalue-\minvalue)))))}%
        \xdef\tempa{\pgfmathresult}%
        \cellcolor{#5!\tempa!#4!#6}\value%
    }}%
}

\definecolor{LightGray}{gray}{0.95} 

\newcommand{\mseCell}[1]{\gradientcell{#1}{0.008929502055014651}{0.2884}{CornflowerBlue}{LightGray}{70}}
\newcommand{\klCell}[1]{\gradientcell{#1}{0.06992975427732642}{4.1397}{LimeGreen}{LightGray}{70}}
\newcommand{\corrCell}[1]{\gradientcell{#1}{0.6354956511466071}{2.092461179611946}{LightGray}{RedOrange}{70}}
\newcommand{\normCell}[1]{\gradientcell{#1}{0.6707819335449955}{3.322553274423951}{LightGray}{Goldenrod}{70}}

\newcommand{\negcorr}[1]{\gradientcell{#1}{-0.26}{0.0}{MidnightBlue}{LightGray}{60}}
\newcommand{\poscorr}[1]{\gradientcell{#1}{0.}{0.26}{LightGray}{BrickRed}{60}}

\begin{document}
\maketitle

\begin{abstract}%
Recent studies evaluate the value orientation of large language models (LLMs) using adapted social surveys, typically by prompting models with survey questions and comparing their responses to average human responses.
This paper identifies limitations in this methodology that, depending on the exact setup, can lead to both underestimating and overestimating the similarity of value orientation.
Using the World Value Survey in three languages across five countries, we demonstrate that prompting methods (direct vs. chain-of-thought) and decoding strategies (greedy vs. sampling) significantly affect results.
To assess the interaction between answers, we introduce a novel metric, self-correlation distance.
This metric measures whether LLMs maintain consistent relationships between answers across different questions, as humans do.
This indicates that even a high average agreement with human data, when considering LLM responses independently, does not guarantee structural alignment in responses.
Additionally, we reveal a weak correlation between two common evaluation metrics, mean-squared distance and KL divergence, which assume that survey answers are independent of each other.
For future research, we recommend CoT prompting, sampling-based decoding with dozens of samples, and robust analysis using multiple metrics, including self-correlation distance.
\end{abstract}

\section{Introduction}

Evaluating the values expressed in texts generated by Large Language Models (LLMs) is crucial for shaping public perception, informing policy, and ensuring the ethical use of AI.
A common evaluation method involves prompting LLMs with questions from standardized surveys and comparing their responses to human answers or calibrated scores (see Table~\ref{tab:related} for a comprehensive list of related work).

This approach has been criticized for its inconsistency with open-ended generation \citep{wright-etal-2024-llm} and sensitivity to prompt formulation \citep{rottger-etal-2024-political,motoki2024more}.
We address a related issue: the reliability of measuring similarity between model and human responses, particularly in how answers to different questions correlate with one another.

We examine this framework using (1) direct vs. Chain-of-Thought (CoT) prompting, (2) greedy decoding vs. nucleus sampling, and (3) three similarity metrics: mean squared difference, KL divergence, and our novel self-correlation distance. Unlike previous metrics that estimate average alignment by treating survey questions in isolation, our metric accounts for interactions among survey answers.

Using LLaMA~3, Mistral 2, EuroLLM, and Qwen~2.5, we compare model responses in the World Value Survey (WVS; \citealp{haerpfer2020world}) with opinions from selected countries.
We find that choices in prompting, decoding, and metrics yield different conclusions.
For example, greedy decoding deviates from typical results obtained with nucleus sampling, while short categorical answers underestimate alignment compared to CoT prompting.
The self-correlation distance indicates that, despite a high average alignment with survey data, different correlation patterns reveal potential overgeneralization. Metrics that treat answers independently can thus overestimate alignment.

In this paper, we first review studies that compare LLM-generated answers with population survey responses (Section~\ref{sec:related}). We then describe our experiments (Section~\ref{sec:experiments}), including the models we use, the prompts we employ, and the inference algorithm we employ. We describe the metrics we use, including the newly introduced self-correlation distance. Section~\ref{sec:results} presents the results, and Section~\ref{sec:conclusions} concludes the paper.

\begin{table*}

\footnotesize\centering
\setlength\tabcolsep{4.1pt}
\begin{tabular}{l cccccc}
\toprule
 & Gold survey data & \begin{minipage}{1cm}\centering Open-source models\end{minipage} & \begin{minipage}{1cm}\centering Per\-sona\end{minipage} & \begin{minipage}{3cm}\centering Decoding \\ (\# samples)\end{minipage} & \begin{minipage}{1cm}\centering Cate\-gorial output\end{minipage} & \begin{minipage}{1.2cm}\centering Compa\-rison w/ humans\end{minipage} \\

\midrule

\citet{santurkar2023opinions} & Custom & No & Yes & Logits & Yes & WD \\
\citet{cao-etal-2023-assessing} & \citet{hofstede1984culture} & No & No & Sampling (1) & Yes & Accuracy \\
\citet{feng-etal-2023-pretraining} & Political compass & Yes & No & 10-best tokens & No & --- \\
\citet{olmedo2023questioning} & \citet{mather2005american} & Yes & No & Logits & Yes & KL \\
\citet{scherrer2023evaluating} & Custom & Yes & No & Full samp. (5/10) & No & --- \\
\citet{sanders2023demonstrations} & Custom & No & Yes & Sampling? (100) & Yes & WD \\
\citet{benkler2023assessing} & WVS & No & Yes & Nucleus samp. (1) & No & --- \\
\citet{tao2024cultural} & WVS, EVS & No & Yes & Greedy & Yes & MSD \\
\citet{durmus2024towards} & Custom & No & Yes & Logits & Yes & JSD \\
\citet{ceron-etal-2024-beyond} & Custom & Yes & Yes &  Nucleus samp. (30) & Yes & --- \\
\citet{nunes2024large} & MFQ, MFV & Yes & No & Sampling? (1) & Yes & --- \\
\citet{vida2024decoding} & \citet{awad2018moral} & Yes & No & Sampling? (1) & Yes & Accuracy \\
\citet{xu2024self} & WVS & Yes & Yes & Nucleus samp. (1) & Yes & Norm. MSD \\
\citet{liu2024measuring} & \citet{pew2018religious} & Yes & No & Greedy & Yes & --- \\
\citet{kim2024exploring} & WVS & Yes & No & Logits & Yes & Pearson \\
\citet{aksoy2024morality} & MFQ & Yes & No & Sampling? (100) & Yes & --- \\

\citet{shen2024valuecompass} & \citet{schwartz1992universals} & Yes & No & Sampling? (10) & Yes & L1 \\

\citet{sukiennik2024evaluation} & \citet{hofstede1984culture} & Yes & No & Greedy & Yes & Norm. L1 \\
\citet{wang2024performance} & WVS & No & Yes & Sampling (100) & Yes & MSD \\
\citet{martinez2024cultural} & WVS & No & No & Not specified & Yes & Accuracy \\
\citet{chen2025prompt} & WVS & Yes & No & Logits & Yes & Accuracy \\
\citet{gurgurov2025multilingual} & Political compass & Yes & No & Nucleus sampling (1) & Yes & --- \\

\citet{bulte2025llms} & \citep{hofstede1984culture} + WVS & Yes & No & Nucleus sampling (6) & Yes & Accuracy \\

\citet{costa2025moral} & MFQ & Yes & Yes & Sampling? (1) & Yes & --- \\

\citet{atari2023which} & WVS & No & No & Sampling? (100) & Yes & Fixation ind.\\ %\citet{muthukrishna2020beyond} \\

% https://arxiv.org/pdf/2512.12488v1 Hofstede

\bottomrule
\end{tabular}

\vspace{1pt}

\emph{Gold survey data}: WVS = World Value Survey \citep{haerpfer2020world}, EVS = European Value Study \citep{evs}, \\ MFQ = Moral Foundation Questionnaire \citep{graham2011MFQ1}, MFV = Moral Foundation Vignettes \citep{clifford2015moral}

\vspace{1pt}

\emph{Comparison methods}: WD = Wasserstein Distance, Acc. = Accuracy, KL = Kullback-Leibler Divergence, \\ JSD = Jensen-Shannon Distance, MSD = Mean Squared Difference

\caption{Overview of publications using standardized surveys to evaluate values in LLMs.}\label{tab:related}

\end{table*}

% ======================================================================
\section{LLMs and Surveys}\label{sec:related}
% ======================================================================

Many studies examine the values encoded in language models, focusing on moral, cultural, and political biases. These often rely on surveys designed for human respondents (see Table~\ref{tab:related}). They cover topics ranging from general ethics to specific domains, such as political ideologies \citep{feng-etal-2023-pretraining}, autonomous vehicle ethics \citep{vida2024decoding}, and religious values \citep{liu2024measuring}.

Model responses are typically evaluated using two approaches: (1) evaluation keys, as in frameworks like the Political Compass or Moral Foundations Questionnaire (\citealp{graham2011MFQ1}; MFQ), or (2) comparisons with human population data.

Most studies prompt models to produce categorical answers, such as selecting an option or providing a score. These responses are compared to population data using metrics like KL Divergence \citep{olmedo2023questioning}, Jensen-Shannon Divergence \citep{durmus2024towards}, or Wasserstein Distance \citep{santurkar2023opinions,sanders2023demonstrations}. However, these methods often focus on single-token generation, which limits their generalization to longer text samples. Some evaluations use greedy decoding or a single sampled response, with sampling details often unspecified. Persona probing (i.e., specifying demographic traits for models to emulate) is also common but typically focuses on English outputs, leaving biases in other languages underexplored.

Several recent studies have extended survey-based evaluation methodologies. \citet{sukiennik2024evaluation} conducted the first large-scale evaluation of cultural alignment across 20 countries and 10 LLMs, using Hofstede's Cultural Values Questionnaire. The results found that models generally represent a moderate cultural middle ground, with the United States showing the best alignment. \citet{wang2024performance} used WVS data to evaluate ChatGPT's public opinion simulation capabilities, revealing significant performance disparities favoring Western, English-speaking nations and demographic biases across gender, ethnicity, and social class. \citet{martinez2024cultural} demonstrated that 44\% of GPT-4o's ability to reflect societal values correlates with digital resource availability in a society's primary language, with error rates in low-resource languages exceeding those in high-resource languages by a factor of five. Most recently, \citet{chen2025prompt} extended bias research in LLM survey responses using WVS data, testing perturbations in answer and question phrasing across multiple models and finding significant sensitivity to prompt variations that mirror human response biases.

\citet{rottger-etal-2024-political,wang-etal-2024-answer-c} and \citet{moore-etal-2024-large} highlight how prompt formulations, such as multiple-choice setups, significantly influence model outputs and score robustness. While their work examines the robustness of the prompt with respect to evaluation keys (such as the Political Compass), it does not address the methodological aspects of comparing model outputs to population survey data. 
Also, as far as we know, all previous work treats survey responses as independent and disregards correlations between questions, an issue we address by introducing the self-correlation distance.

Previous work also evaluated generation consistency \citep{kumar-joshi-2022-striking,bonagiri-etal-2024-sage}. However, it focuses on cases with a well-specified ground truth. We are interested in a slightly different type of consistency: Statements like people who tend to say $A$, also tend to say $B$ to some extent. In this paper, we introduce a metric to measure the extent to which LLMs capture these tendencies, independent of the actual content.

Recent work has also examined cultural adaptation methods using survey data. \citet{adilazuarda2024surveys} found that WVS-based training can lead to cultural homogenization and undermine factual knowledge, and introduced a cultural distinctiveness metric that complements existing evaluation approaches. Their findings that survey data alone may be insufficient for cultural adaptation align with our observations about the limitations of current evaluation methodologies.

% ======================================================================
\section{Experiments}\label{sec:experiments}
% ======================================================================

Following several previous studies \citep{benkler2023assessing,tao2024cultural,kim2024exploring}, we prompt LLMs with World Value Survey (Round~7, version~5.0) questions and compare their answers with human data using three evaluation methods.
WVS is a global research project that has comprehensively explored people's values and beliefs since 1981.
The World Value Survey covers 55 countries and 80 languages, making it likely the most comprehensive standardized resource for comparing values in LLM outputs with the human population across languages and cultures.
In this work, we focus on evaluation metrics and conduct experiments only on a small subset.

As in related work, we simulate the survey using an LLM and compare the results with those obtained from the human population in the respective countries. Since the answers to all questions are integers, depending on the evaluation metric, we either use the average answers or the categorical distributions of answers as the ground truth for comparison. For the correlation study, we use responses from individual respondents and compute how responses to individual questions correlate with each other.

The source code to replicate the experiments is available at \url{https://github.com/jlibovicky/llm-survey-eval}.

\subsection{Model Prompting}\label{ssec:prompting}

\paragraph{Questionnaire design.}

WVS is not a self-assessment questionnaire.
Interlocutors interview the subjects and, based on their answers, record integer scores for each question, most often on a scale from 1 to 10, indicating the extent to which they agree with a statement.

We use general, non-personalized formulations of the questions, i.e., we exclude questions about income, health, or personal experiences, which would likely be rejected by the models. We reformulated the questions to contain more general, non-personal statements (e.g., replacing ``your life'' with ``human life'') to further reduce rejections. After excluding questions that were not in all language versions, 143 prompts remained. Prompts were created in English, machine-translated into German and Czech using Google Translate, and then manually post-edited with reference to official WVS translations by native speakers. The questionnaire was administered in a single conversation session to allow evaluation of answer consistency (see examples in Appendix~\ref{sec:appendix}).

\paragraph{Scores vs. Chain-of-Thought.}

We compare two prompt types: direct numeric answers and chain-of-thought prompts \citep{cot}, where justification precedes the answer. We posit that chain-of-thought is closer to real-world LLM use, as chat-like user interactions typically involve longer generations than a single categorical output, e.g., in interactive sessions.

\paragraph{Greedy vs. Sampling.}

Studies often use greedy decoding for its efficiency and for producing a deterministic output. It approximates the most probable sequence but may not yield typical responses, as the probability mass is distributed across similar sequences \citep{eikema-aziz-2020-map, wiher-etal-2022-decoding}. Because of this, and since sampling is more common in practice, we compare greedy decoding with nucleus sampling (nucleus 0.9, temperature 0.7), including an estimation of how many samples are needed to obtain a stable result for the given metrics. Following \citet{andreas-2022-language}, who argues that language models should be treated as ensembles of different multiple agents, and \citet{lederman-mahowald-2024-language}, who argue that language models are compressed libraries, we assume that sampling one conversation session might correspond to one agent within the language model in the agent metaphor and retrieving one set of world knowledge from the library in the library metaphor. Therefore, we treat the conversation sessions as comparable to individual respondents in the survey.

\subsection{Evaluation}\label{ssec:evaluation}

We evaluate model outputs against data from the USA, UK, Czechia, and Germany, where the prompt languages are spoken. Iran and China are included as culturally distinct cases, a sanity check for metric validation.\footnote{The number of WVS participants in the respective countries was: USA: 2,596, UK: 2,609, Czechia: 1,200, Germany: 1,528, Iran: 1,499, China: 3,036.} Based on the results of previous studies, we expect that due to the prevalence of English data, models will tend to better align with Western countries. We use three metrics: Mean Squared Difference (MSD), Kullback-Leibler Divergence (KLD), and a novel self-correlation distance.

\paragraph{Models.}

We use four instruction-tuned models: LLaMA~3 8B Instruct \citep{llama3} and Mistral v0.1 7B Instruct \citep{mistral7b}, EuroLLM 9B Instruct \citep{martins2024eurollm}, and Qwen 2.5 7B Instruct \citep{yang2024qwen}. This selection includes both general-purpose models (LLaMA~3 from the USA, Mistral from France) and models with specific regional focuses (EuroLLM for European languages, Qwen developed in China for multilingual applications), allowing us to examine how model design influences value alignment across cultures.

\paragraph{Mean Square Difference.}

Most WVS questions have numerical answers on a 1-to-10 scale or lower. We scale the answer to the 0--1 interval, compute the squared differences between the scores sampled from the model and human population averages, and average them over all questions.

\paragraph{KL Divergence.}

We treat the model and human answers as categorical distributions. For the model outputs, we normalize the distribution over model runs. For the survey data, we normalize over the participants. We compute the Kullback-Leibler divergence between the sampled model answer distribution and the distribution of answers in the human population.

\paragraph{Self-Correlation Distance.}

The previous metrics fail to account for the fact that questionnaire responses often correlate with one another due to underlying consistency in values and opinions, yet all questions are treated as conditionally independent. This assumption is not realistic.
Value opinions often come in bundles and follow patterns that may differ across various societies. Simple examples might include people who believe that religion should play a stronger role in society being more likely to say that mothers of young children should stay at home with their children, or that individuals concerned about social justice are often also concerned about the environment.
These correlations between individual respondents' answers capture second-order patterns, relationships between answers, that are not apparent when comparing individual answers alone, as with MSE or KL-Divergence.

To analyze this, we compute self-correlation matrices that measure the Pearson correlation between all pairs of questionnaire responses. In the survey data, we calculate the correlation between responses to individual questions across participants. In the LLM case, we calculate the correlation between answers across model runs. This gives a matrix with all question pairs. Using the Frobenius norm, we quantify how internally consistent or ``principled'' the answers are. High norm means that the absolute value of the correlations tends to be high, whereas a norm close to zero means that questions in the survey are more independent of each other.

Additionally, we compare the self-correlation matrices of model outputs and human responses using the Frobenius norm. This metric allows us to evaluate whether the underlying structure of the answers aligns between models and humans, going beyond simple agreement on individual responses taken independently.

% ======================================================================
\section{Results}\label{sec:results}
% ======================================================================

\paragraph{Comparing MSD, KLD, and self-correlation distance.}

The results comparing model responses to human surveys are presented in Table~\ref{tab:usa_compare} for the USA, with additional results for other countries and languages provided in Table~\ref{tab:model_compare} in the Appendix. Within languages and countries, results follow similar trends across setups.

%\paragarph{Contextualizing the metrics.}
To interpret our results, we first establish baseline values from human populations. Country-level comparisons in the WVS yield MSD values ranging from 0.009 (USA-UK) to 0.069 (Germany-Iran), with Western countries showing differences of 0.009--0.024 (see Table~\ref{tab:country_compare} in the Appendix). KLD between countries ranges from 0.07 (USA-UK) to 0.44 (Germany-Iran), with Western countries showing 0.07--0.22. Self-correlation distances between human populations range from 0.64 (China-Iran) to 0.95 (USA-Iran), with typical values between 0.79 and 0.92.

%\textbf{Surface-level alignment varies dramatically by setup.}
Using MSD and KLD metrics, which treat answers independently, we observe substantial variation across prompting and decoding strategies. With CoT prompting and sampling, Mistral 2 achieves remarkably low MSD (0.022) and KLD (0.26) for USA data, comparable to differences between Western countries. However, the same model with greedy decoding and CoT prompting shows drastically worse alignment (MSD=0.188), exceeding even USA-Iran differences. LLaMA~3 shows more moderate values (MSD=0.059, KLD=1.47 for CoT+Sampling), falling between Western and cross-cultural differences. EuroLLM exhibits the poorest alignment with direct prompting and greedy decoding (MSD=0.165--0.284), though sampling decreases the distance between model and population substantially. Qwen demonstrates relatively stable performance across setups (MSD=0.041--0.199).

%\textbf{Structural alignment tells a different story.}
The self-correlation distance metric reveals a paradox: setups that achieve the best surface-level alignment often exhibit the poorest structural alignment. Mistral~2, with CoT and sampling, despite having the lowest MSD (0.022) and KLD (0.26), exhibits a self-correlation distance of 1.62, which is far greater than the distances between any human populations (0.64--0.95). Its correlation norm (2.80) is also substantially higher than human values (${\sim}$1.66), which indicates overly rigid response patterns. In contrast, LLaMA~3 with the same setup shows better structural alignment (self-correlation distance=1.29, correlation norm=1.70) despite worse surface metrics (MSD=0.059, KLD=1.47). Qwen, with score-only prompts and sampling, produces the most structured responses (correlation norm=3.33, self-correlation distance=2.13) with an even greater departure from human response variability.

%\textbf{Greedy decoding is problematic.}
Across all models and prompting strategies, greedy decoding consistently underestimates alignment when measured with MSD. For instance, comparing LLaMA~3 with score-only prompts, greedy decoding yields MSD=0.098, versus 0.073 with sampling, roughly a one-third increase. The disparity is even more pronounced for KLD, where sampling values are often 2--3$\times$ higher than greedy decoding, indicating that greedy decoding captures only a narrow slice of the probability distribution reachable by common decoding algorithms.

%\textbf{Direct prompting systematically differs from CoT.}
The effect of CoT prompting varies across models and decoding strategies. For LLaMA~3 with sampling, CoT improves alignment (MSD decreases from 0.073 to 0.059). For Mistral~2 with sampling, CoT dramatically improves both MSD (0.041$\rightarrow$0.022) and KLD (0.77$\rightarrow$0.26). However, with greedy decoding, CoT can worsen results: Mistral~2's MSD increases from 0.094 to 0.188. This suggests that CoT prompting is beneficial primarily when combined with sampling-based decoding, likely because it provides more stable generation that better leverages the sampling strategy.

%\textbf{Metric correlations differ substantially by model.}

Table~\ref{tab:metric_correlation} shows moderate overall correlation between MSD and KLD (Pearson=0.465, Spearman=0.717), but Table~\ref{tab:metric_correlation_breakdown} reveals this masks dramatic model-specific differences. Mistral~2 shows a very strong correlation (Pearson=0.832, Spearman=0.925), meaning MSD and KLD largely agree on what constitutes good alignment for this model. LLaMA~3, however, shows weak correlation (Pearson=0.276, Spearman=0.389), indicating that these metrics measure somewhat different aspects of value alignment for this model. Critically, self-correlation distance is negatively correlated with both MSD (Pearson=$-0.389$) and KLD (Pearson=$-0.083$), confirming that surface-level and structural alignment are different, and sometimes opposing, qualities.

\begin{table}[t]
\footnotesize\centering

\begin{tabular}{l@{\hskip 8pt} ll c c c c}
\toprule
\rotatebox[origin=c]{90}{Model} & \begin{minipage}{.8cm}Prompt type\end{minipage} & \begin{minipage}{.2cm}De\-co\-de\end{minipage}& 
\begin{minipage}{.6cm}\centering MSD\end{minipage} &
\begin{minipage}{.6cm}\centering KLD\end{minipage} &
\begin{minipage}{.6cm}\centering Corr. norm\end{minipage} &
\begin{minipage}{.6cm}\centering Self-corr. dist.\end{minipage}
\\ \midrule

\multirow{4}{*}{\rotatebox[origin=c]{90}{LLaMA~3~}}
& \multirow{2}{*}{\begin{minipage}{.8cm}Score only\end{minipage}} & Gr. &
\mseCell{.098} & \klCell{1.71}  \\
& & Spl. &
\mseCell{.073} & \klCell{2.99} & \normCell{0.90} & \corrCell{1.26} \\ \cmidrule{2-7}
& \multirow{2}{*}{CoT} & Gr. &
\mseCell{.088} & \klCell{1.68} \\
& & Spl. & 
\mseCell{.059} & \klCell{1.47} & \normCell{1.70} & \corrCell{1.29} \\
\midrule
\multirow{4}{*}{\rotatebox[origin=c]{90}{Mistral2~}}
& \multirow{2}{*}{\begin{minipage}{.8cm}Score only\end{minipage}} & Gr. &
\mseCell{.094} & \klCell{1.80}  \\
& & Spl. &
\mseCell{.041} & \klCell{0.77} & \normCell{1.77} & \corrCell{1.17} \\ \cmidrule{2-7}
& \multirow{2}{*}{CoT} & Gr. &
\mseCell{.188} & \klCell{1.95} \\
& & Spl. & 
\mseCell{.022} & \klCell{0.26} & \normCell{2.80} & \corrCell{1.62} \\
\midrule
\multirow{4}{*}{\rotatebox[origin=c]{90}{EuroLLM~}}
& \multirow{2}{*}{\begin{minipage}{.8cm}Score only\end{minipage}} & Gr. &
\mseCell{.165} & \klCell{1.91}  \\
& & Spl. &
\mseCell{.059} & \klCell{0.72} & \normCell{2.55} & \corrCell{1.56} \\ \cmidrule{2-7}
& \multirow{2}{*}{CoT} & Gr. &
\mseCell{.130} & \klCell{1.76} \\
& & Spl. & 
\mseCell{.125} & \klCell{0.97} & \normCell{2.97} & \corrCell{1.74} \\
\midrule
\multirow{4}{*}{\rotatebox[origin=c]{90}{Qwen~}}
& \multirow{2}{*}{\begin{minipage}{.8cm}Score only\end{minipage}} & Gr. &
\mseCell{.082} & \klCell{1.66}  \\
& & Spl. &
\mseCell{.074} & \klCell{1.27} & \normCell{3.33} & \corrCell{2.13} \\ \cmidrule{2-7}
& \multirow{2}{*}{CoT} & Gr. &
\mseCell{.199} & \klCell{1.60} \\
& & Spl. & 
\mseCell{.062} & \klCell{1.13} & \normCell{1.52} & \corrCell{1.25} \\

\bottomrule
\end{tabular}

\caption{A comparison of model outputs with different prompting strategies (Score-only, CoT: Chain of Thought), decoding method (Gr.: greedy, Spl.: sampling). For other countries and languages, see Table~\ref{tab:model_compare} in the Appendix.}\label{tab:usa_compare}
\end{table}

\begin{table}[t]

\footnotesize\centering

\begin{tabular}{l ccc}
\toprule
 & MSD & KLD & CorrD \\
\midrule
MSD & ---  & \hphantom{-}.465 & -.389 \\
KLD & \hphantom{-}.717 & --- & -.083 \\
CorrD & -.374 & -.064 & ---  \\
\bottomrule
\end{tabular}

\caption{Correlation of the metrics (Pearson above the diagonal, Spearman under the diagonal) for both models over all languages and countries.}\label{tab:metric_correlation}

\end{table}

\begin{table}[t]
\footnotesize\centering
\setlength\tabcolsep{4.3pt}

\begin{tabular}{l@{\hskip 7pt} ll cc@{\hskip 10pt} cc@{\hskip 10pt} cc}
\toprule
\rotatebox[origin=c]{90}{Model} & \begin{minipage}{.8cm}Prompt type\end{minipage} & \begin{minipage}{.6cm}Deco\-de\end{minipage}&
\multicolumn{2}{c}{\begin{minipage}{1.0cm}\centering MSD\end{minipage}} &
\multicolumn{2}{c}{\begin{minipage}{1.0cm}\centering KLD\end{minipage}} &
\multicolumn{2}{c}{\begin{minipage}{1.0cm}\centering Self-corr. dist.\end{minipage}}
\\ \midrule

\multirow{4}{*}{\rotatebox[origin=c]{90}{LLaMA~3~}}
& \multirow{2}{*}{\begin{minipage}{.8cm}Score only\end{minipage}} & Gr. & \negcorr{-.19} & \poscorr{.18} & \negcorr{-.07} & \poscorr{.21} &  \\
& & Spl. & \negcorr{-.08} & \poscorr{.24} & \poscorr{.08} & \poscorr{.24} & \poscorr{.02} & \negcorr{-.25} \\ \cmidrule{2-6}
& \multirow{2}{*}{CoT} & Gr. & \negcorr{-.12} & \poscorr{.26} & \negcorr{-.29} & \poscorr{.21} & \\
& & Spl. &  \negcorr{-.03} & \poscorr{.34} & \negcorr{-.17} & \poscorr{.12} & \negcorr{-.12} & \negcorr{-.28} \\
\midrule
\multirow{4}{*}{\rotatebox[origin=c]{90}{Mistral2~}}
& \multirow{2}{*}{\begin{minipage}{.8cm}Score only\end{minipage}} & Gr. & \poscorr{.26} & \poscorr{.23} & \poscorr{.25} & \poscorr{.36} &  \\
& & Spl. & \poscorr{.01} & \poscorr{.33} & \negcorr{-.08} & \poscorr{.17} & \poscorr{.07} & \poscorr{.01} \\ \cmidrule{2-6}
& \multirow{2}{*}{CoT} & Gr. & \negcorr{-.16} & \poscorr{.08} & \negcorr{-.01} & \poscorr{.13} & \\
& & Spl. &  \poscorr{.24} & \poscorr{.26} & \poscorr{.21} & \poscorr{.20} & \poscorr{.19} & \poscorr{.21} \\
\midrule
\multirow{4}{*}{\rotatebox[origin=c]{90}{EuroLLM~}}
& \multirow{2}{*}{\begin{minipage}{.8cm}Score only\end{minipage}} & Gr. & \poscorr{.37} & \poscorr{.16} & \poscorr{.30} & \poscorr{.08} &  \\
& & Spl. & \poscorr{.24} & \negcorr{-.01} & \poscorr{.18} & \poscorr{.07} & \poscorr{.02} & \poscorr{.03} \\ \cmidrule{2-6}
& \multirow{2}{*}{CoT} & Gr. & \poscorr{.28} & \poscorr{.12} & \negcorr{-.18} & \negcorr{-.16} & \\
& & Spl. &  \poscorr{.07} & \negcorr{-.26} & \poscorr{.15} & \negcorr{-.15} & \negcorr{-.17} & \negcorr{-.00} \\
\midrule
\multirow{4}{*}{\rotatebox[origin=c]{90}{Qwen~}}
& \multirow{2}{*}{\begin{minipage}{.8cm}Score only\end{minipage}} & Gr. & \poscorr{.46} & \poscorr{.45} & \negcorr{-.20} & \negcorr{-.07} &  \\
& & Spl. & \negcorr{-.26} & \negcorr{-.00} & \negcorr{-.14} & \poscorr{.03} & \negcorr{-.17} & \negcorr{-.15} \\ \cmidrule{2-6}
& \multirow{2}{*}{CoT} & Gr. & \negcorr{-.12} & \poscorr{.11} & \negcorr{-.10} & \poscorr{.03} & \\
& & Spl. &  \negcorr{-.05} & \poscorr{.28} & \negcorr{-.12} & \poscorr{.09} & \poscorr{.03} & \negcorr{-.16} \\

\bottomrule
\end{tabular}

\begin{tikzpicture}
  \pgfdeclarehorizontalshading{minusshading}{100bp}
    {color(0bp)=(MidnightBlue); color(100bp)=(LightGray)}
  \pgfdeclarehorizontalshading{plusshading}{100bp}
    {color(0bp)=(LightGray); color(100bp)=(BrickRed)}

    % Draw the node with the gradient
    \node[shade, shading=minusshading, minimum width=1.5cm, minimum height=0.3cm, opacity=0.7] (rec1) {};
    \node[shade, shading=plusshading, minimum width=1.5cm, minimum height=0.3cm, opacity=0.7, right=-1pt of rec1] (rec2) {};
    \node[yshift=-1pt, left=0pt of rec1] {\footnotesize\vphantom{g} Color scale from};
    \node[yshift=-1pt, left=-13pt of rec1] {\footnotesize\vphantom{g} -.3};
    \node[yshift=-1pt, left=-17pt of rec2] {\footnotesize\vphantom{C} through 0};
    \node[yshift=-1pt, right=-21pt of rec2] {\footnotesize\vphantom{g} to .3};
    
\end{tikzpicture}

\caption{Point-Biserial Correlation of the alignment metrics with matching country and language as a binary indicator variable. The left part shows the correlation only within Western countries; the right part also includes Iran and China.}\label{tab:match_countires}

\end{table}

\paragraph{Cross-lingual and cross-cultural patterns.}

Table~\ref{tab:match_countires} summarizes how language-country matching affects alignment metrics. The correlations are generally weak, but patterns emerge when culturally distant countries (Iran and China) are included. Most models exhibit positive correlations between matching conditions and alignment metrics, suggesting that prompts in English, Czech, and German align more closely with Western surveys than with those from Iran or China. However, the strength of this effect varies: EuroLLM shows the strongest language-country matching effects (up to 0.37), likely reflecting its European-centric training, while Qwen exhibits more uniform performance across language-country combinations, consistent with its multilingual design.

Notably, the correlation patterns differ substantially across prompting and decoding strategies. Mistral~2 with CoT prompting and nucleus sampling shows the most consistent positive correlations with both language matching (0.24 for MSD, 0.21 for KLD) and the inclusion of distant cultures (0.26 for MSD, 0.20 for KLD). This suggests that this setup not only achieves the best average alignment but also shows more predictable cross-cultural patterns.

\begin{figure}[t]
    \centering
    \includegraphics[width=\columnwidth]{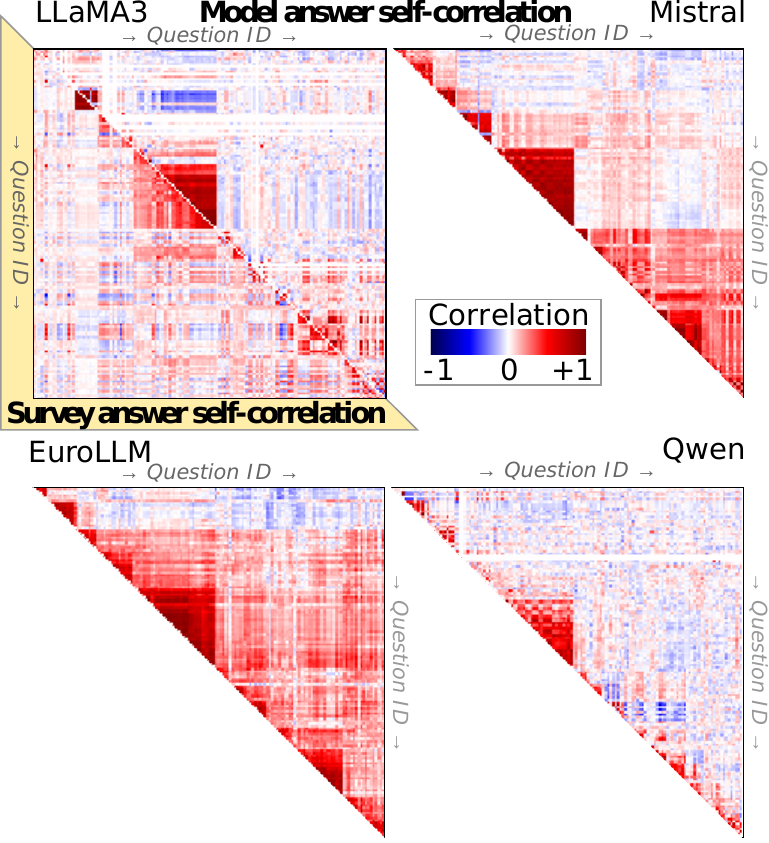}
    \caption{Correlation patterns between human answers in the USA (under the diagonal) and between answers of the LLaMA~3 and Mistral 2 models in English (above the diagonal).}
    \label{fig:self_correlation}
\end{figure}

\begin{figure}[t]
    \centering
    \includegraphics[width=\linewidth]{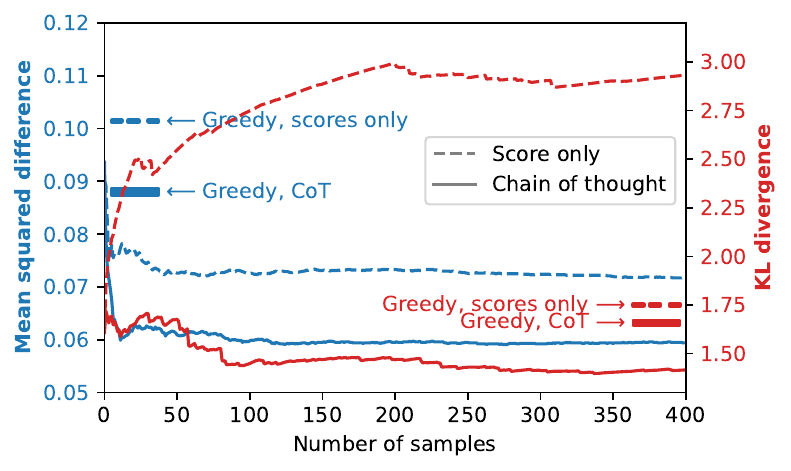}
    \caption{\textcolor{pltblue}{Mean-squared difference} and \textcolor{pltgreen}{KL-divergence} of LLaMA~3 answers when compared to the USA data of the World Value Survey. It compares the greedy decoding and sampling from the model.}
    \label{fig:convergence}
\end{figure}

\paragraph{Self-correlation patterns.}

Figure~\ref{fig:self_correlation} visualizes the self-correlation matrices, revealing systematic differences in how models structure their responses compared to humans.
The visualization shows three major patterns: (1) Block diagonal structures indicate question clusters with strong internal correlations. (2) The intensity of colors shows correlation strength: models produce darker, more saturated colors than humans. (3) Off-diagonal patterns reveal how different value domains relate;  models show simpler, more predictable patterns than the human answers.

All models show overly consistent responses in sections about cultural identity and national pride (visible as darker red blocks along the diagonal). Mistral~2 displays particularly strong self-correlation in questions about social and political attitudes, far exceeding human patterns. These visualizations confirm that, while models may match human averages on individual questions, their internal response structures systematically diverge from human response patterns.

The quantitative self-correlation distances reinforce this finding: model-to-human distances (1.1--2.1 across setups) consistently exceed human-to-human distances (0.64--0.95), indicating that all tested models impose more rigid correlation structures than exist in human populations. This suggests that models generate ``principled'' but overly simplistic response patterns, potentially missing the nuanced and sometimes inconsistent nature of human value systems.

\paragraph{Number of sampled responses.}

Figure~\ref{fig:convergence} demonstrates how metric estimates stabilize with increasing sample size. For MSD, approximately 100 samples are needed for stable estimates, with values converging to within 0.005 of the final estimate. KLD requires more samples, several hundred for full stability. CoT prompting produces more stable estimates with fewer samples compared to direct prompts, likely because the reasoning process provides additional structure that reduces sampling variance.
Importantly, greedy decoding produces estimates that deviate significantly from the sampling-based norm across both metrics and all sample sizes. This confirms that greedy decoding systematically misestimates both average alignment and response distributions. Most prior studies (Table~\ref{tab:related}) used far fewer samples (typically 1--10), suggesting their conclusions may have differed substantially with more comprehensive sampling, particularly for KLD estimates.

\paragraph{Model-specific behaviors.}

The four tested models show distinct patterns. Mistral~2 achieves the best surface-level scores but at the cost of the highest structural rigidity, suggesting it may be overfitting to typical responses. LLaMA~3 shows the most balanced profile across metrics. EuroLLM's strong language-matching effects (correlation of up to 0.37) reflect its European-centric training data. Qwen, despite being developed in China, shows relatively uniform cross-lingual performance but produces highly structured responses (correlation norms up to 3.33).

% ======================================================================
\section{Conclusions}\label{sec:conclusions}
% ======================================================================

This study examined the impact of decoding strategies and evaluation metrics on comparisons of LLM responses to population survey data, using the World Value Survey as a case study across three languages and six countries.

We found that setups closely mirroring real-world LLM usage, specifically, Chain-of-Thought prompting with sampling-based decoding, achieve the best alignment with survey data. \textbf{Prior work relying on direct prompts and greedy decoding may underestimate average alignment when evaluating answers independently.}

To address gaps in current evaluations, we introduced the self-correlation distance, a novel metric that captures consistency and interaction between answers. Unlike traditional metrics such as MSD and KL Divergence, the self-correlation distance showed that high scores in some setups indicate a lower diversity of responses than in the human population. The high average numbers are achieved at the cost of generating typical cases, resulting in overly structured responses rather than an accurate reflection of survey variability, especially in social and political topics. \textbf{This shows that metrics treating answers independently overestimate the alignment.} Discrepancies between metrics, such as LLaMA~3's low correlation between MSD and KLD (Pearson=0.276), underscore the importance of multi-metric evaluations.

For future research, we recommend using CoT prompting, nucleus sampling with at least 100 samples to achieve stable estimates, and a multi-metric approach that incorporates our proposed self-correlation distance to capture both surface-level and structural alignment.

% ======================================================================
\section*{Limitations}
% ======================================================================

The assumption behind all surveys is that the answers provided by the survey subjects reflect their actual behavior in the real world. For example, a man might claim that he believes men should do a fair share of invisible household labor and will vote for parties with a compatible election program. With LLMs, there is no such guarantee. LLMs might advocate for certain values when prompted to generate text directly related to those values, but still generate texts with underlying values that do not align with the survey answer. We do not challenge this assumption in this paper. We are not aware of any existing methodology that would approach this challenge.

Comparing answers across countries based on answer distributions is a simplifying assumption. Demographic factors other than nationality, such as age or economic status, may also play a role. This study used cross-country distributions to gain insights into LLM evaluation, rather than making claims about LLMs being more representative of certain countries than others. Such claims would require a more detailed methodology.

The primary objective of this study was to compare metrics; therefore, we included only a few variables that could influence the model's behavior. It is possible that different prompt formulations and question order in the questionnaire may also yield slightly different results.

\section*{Acknowledgments}

Many thanks to Jan Hajič Jr. and Dominik Macháček for comments on the draft of this paper.

We used GitHub Copilot when writing our code. When writing the paper, we used Grammarly and Claude to improve grammar and spelling.

This research was supported by the Charles University project PRIMUS/23/SCI/023 and project CZ.02.01.01/00/23\_020/0008518 of the Czech Ministry of Education.

% TODO Do not forget to add the acknowledgments into the camera ready.

%\nocite{*}
\bibliography{anthology,custom}

\begin{thebibliography}{51}
\providecommand{\natexlab}[1]{#1}

\bibitem[{Adilazuarda et~al.(2025)Adilazuarda, Liu, Gurevych, and
  Aji}]{adilazuarda2024surveys}
Muhammad~Farid Adilazuarda, Chen~Cecilia Liu, Iryna Gurevych, and Alham~Fikri
  Aji. 2025.
\newblock \href {https://doi.org/10.48550/ARXIV.2505.16408} {From surveys to
  narratives: Rethinking cultural value adaptation in llms}.
\newblock \emph{CoRR}, abs/2505.16408.

\bibitem[{Aksoy(2024)}]{aksoy2024morality}
Meltem Aksoy. 2024.
\newblock \href {https://doi.org/10.48550/ARXIV.2412.18863} {Whose morality do
  they speak? unraveling cultural bias in multilingual language models}.
\newblock \emph{CoRR}, abs/2412.18863.

\bibitem[{Andreas(2022)}]{andreas-2022-language}
Jacob Andreas. 2022.
\newblock \href {https://doi.org/10.18653/v1/2022.findings-emnlp.423} {Language
  models as agent models}.
\newblock In \emph{Findings of the Association for Computational Linguistics:
  EMNLP 2022}, pages 5769--5779, Abu Dhabi, United Arab Emirates. Association
  for Computational Linguistics.

\bibitem[{Atari et~al.(2023)Atari, Xue, Park, Blasi, and
  Henrich}]{atari2023which}
Mohammad Atari, Mona~J. Xue, Peter~S. Park, Dami{\'a}n~E. Blasi, and Joseph
  Henrich. 2023.
\newblock \href {https://doi.org/10.31234/osf.io/5b26t} {Which humans?}
\newblock \emph{PsyArXiv}.

\bibitem[{Awad et~al.(2018)Awad, Dsouza, Kim, Schulz, Henrich, Shariff,
  Bonnefon, and Rahwan}]{awad2018moral}
Edmond Awad, Sohan Dsouza, Richard Kim, Jonathan Schulz, Joseph Henrich, Azim
  Shariff, Jean-Fran{\c{c}}ois Bonnefon, and Iyad Rahwan. 2018.
\newblock The moral machine experiment.
\newblock \emph{Nature}, 563(7729):59--64.

\bibitem[{Benkler et~al.(2023)Benkler, Mosaphir, Friedman, Smart, and
  Schmer{-}Galunder}]{benkler2023assessing}
Noam Benkler, Drisana Mosaphir, Scott Friedman, Andrew Smart, and Sonja
  Schmer{-}Galunder. 2023.
\newblock \href {https://doi.org/10.48550/ARXIV.2312.10075} {Assessing llms for
  moral value pluralism}.
\newblock \emph{CoRR}, abs/2312.10075.

\bibitem[{Bonagiri et~al.(2024)Bonagiri, Vennam, Govil, Kumaraguru, and
  Gaur}]{bonagiri-etal-2024-sage}
Vamshi~Krishna Bonagiri, Sreeram Vennam, Priyanshul Govil, Ponnurangam
  Kumaraguru, and Manas Gaur. 2024.
\newblock \href {https://aclanthology.org/2024.lrec-main.1243} {{S}a{GE}:
  Evaluating moral consistency in large language models}.
\newblock In \emph{Proceedings of the 2024 Joint International Conference on
  Computational Linguistics, Language Resources and Evaluation (LREC-COLING
  2024)}, pages 14272--14284, Torino, Italia. ELRA and ICCL.

\bibitem[{Bulté and Rigouts(2025)}]{bulte2025llms}
Bram Bulté and Terryn~Ayla Rigouts. 2025.
\newblock \href {https://doi.org/10.1162/COLI.a.583} {Llms and cultural values:
  The impact of prompt language and explicit cultural framing}.
\newblock \emph{Computational Linguistics}, pages 1--85.

\bibitem[{Cao et~al.(2023)Cao, Zhou, Lee, Cabello, Chen, and
  Hershcovich}]{cao-etal-2023-assessing}
Yong Cao, Li~Zhou, Seolhwa Lee, Laura Cabello, Min Chen, and Daniel
  Hershcovich. 2023.
\newblock \href {https://doi.org/10.18653/v1/2023.c3nlp-1.7} {Assessing
  cross-cultural alignment between {C}hat{GPT} and human societies: An
  empirical study}.
\newblock In \emph{Proceedings of the First Workshop on Cross-Cultural
  Considerations in NLP (C3NLP)}, pages 53--67, Dubrovnik, Croatia. Association
  for Computational Linguistics.

\bibitem[{Center(2018)}]{pew2018religious}
Pew~Research Center. 2018.
\newblock The religious typology: A new way to categorize americans by
  religion.

\bibitem[{Ceron et~al.(2024)Ceron, Falk, Bari{\'c}, Nikolaev, and
  Pad{\'o}}]{ceron-etal-2024-beyond}
Tanise Ceron, Neele Falk, Ana Bari{\'c}, Dmitry Nikolaev, and Sebastian
  Pad{\'o}. 2024.
\newblock \href {https://doi.org/10.1162/tacl_a_00710} {Beyond prompt
  brittleness: Evaluating the reliability and consistency of political
  worldviews in {LLM}s}.
\newblock \emph{Transactions of the Association for Computational Linguistics},
  12:1378--1400.

\bibitem[{Clifford et~al.(2015)Clifford, Iyengar, Cabeza, and
  Sinnott-Armstrong}]{clifford2015moral}
Scott Clifford, Vijeth Iyengar, Roberto Cabeza, and Walter Sinnott-Armstrong.
  2015.
\newblock Moral foundations vignettes: A standardized stimulus database of
  scenarios based on moral foundations theory.
\newblock \emph{Behavior research methods}, 47(4):1178--1198.

\bibitem[{Costa et~al.(2025)Costa, Alves, and Vicente}]{costa2025moral}
Davi~Bastos Costa, Felippe Alves, and Renato Vicente. 2025.
\newblock \href {https://doi.org/10.48550/ARXIV.2511.08565} {Moral
  susceptibility and robustness under persona role-play in large language
  models}.
\newblock \emph{CoRR}, abs/2511.08565.

\bibitem[{Dubey et~al.(2024)Dubey, Jauhri, Pandey, Kadian, Al{-}Dahle, Letman,
  Mathur, Schelten, Yang, Fan, Goyal, Hartshorn, Yang, Mitra, Sravankumar,
  Korenev, Hinsvark, Rao, Zhang, Rodriguez, Gregerson, Spataru, Rozi{\`{e}}re,
  Biron, Tang, Chern, Caucheteux, Nayak, Bi, Marra, McConnell, Keller, Touret,
  Wu, Wong, Ferrer, Nikolaidis, Allonsius, Song, Pintz, Livshits, Esiobu,
  Choudhary, Mahajan, Garcia{-}Olano, Perino, Hupkes, Lakomkin, AlBadawy,
  Lobanova, Dinan, Smith, Radenovic, Zhang, Synnaeve, Lee, Anderson, Nail,
  Mialon, Pang, Cucurell, Nguyen, Korevaar, Xu, Touvron, Zarov, Ibarra,
  Kloumann, Misra, Evtimov, Copet, Lee, Geffert, Vranes, Park, Mahadeokar,
  Shah, van~der Linde, Billock, Hong, Lee, Fu, Chi, Huang, Liu, Wang, Yu,
  Bitton, Spisak, Park, Rocca, Johnstun, Saxe, Jia, Alwala, Upasani, Plawiak,
  Li, Heafield, Stone et~al.}]{llama3}
Abhimanyu Dubey, Abhinav Jauhri, Abhinav Pandey, Abhishek Kadian, Ahmad
  Al{-}Dahle, Aiesha Letman, Akhil Mathur, Alan Schelten, Amy Yang, Angela Fan,
  Anirudh Goyal, Anthony Hartshorn, Aobo Yang, Archi Mitra, Archie Sravankumar,
  Artem Korenev, Arthur Hinsvark, Arun Rao, Aston Zhang, Aur{\'{e}}lien
  Rodriguez, Austen Gregerson, Ava Spataru, Baptiste Rozi{\`{e}}re, Bethany
  Biron, Binh Tang, Bobbie Chern, Charlotte Caucheteux, Chaya Nayak, Chloe Bi,
  Chris Marra, Chris McConnell, Christian Keller, Christophe Touret, Chunyang
  Wu, Corinne Wong, Cristian~Canton Ferrer, Cyrus Nikolaidis, Damien Allonsius,
  Daniel Song, Danielle Pintz, Danny Livshits, David Esiobu, Dhruv Choudhary,
  Dhruv Mahajan, Diego Garcia{-}Olano, Diego Perino, Dieuwke Hupkes, Egor
  Lakomkin, Ehab AlBadawy, Elina Lobanova, Emily Dinan, Eric~Michael Smith,
  Filip Radenovic, Frank Zhang, Gabriel Synnaeve, Gabrielle Lee, Georgia~Lewis
  Anderson, Graeme Nail, Gr{\'{e}}goire Mialon, Guan Pang, Guillem Cucurell,
  Hailey Nguyen, Hannah Korevaar, Hu~Xu, Hugo Touvron, Iliyan Zarov,
  Imanol~Arrieta Ibarra, Isabel~M. Kloumann, Ishan Misra, Ivan Evtimov, Jade
  Copet, Jaewon Lee, Jan Geffert, Jana Vranes, Jason Park, Jay Mahadeokar, Jeet
  Shah, Jelmer van~der Linde, Jennifer Billock, Jenny Hong, Jenya Lee, Jeremy
  Fu, Jianfeng Chi, Jianyu Huang, Jiawen Liu, Jie Wang, Jiecao Yu, Joanna
  Bitton, Joe Spisak, Jongsoo Park, Joseph Rocca, Joshua Johnstun, Joshua Saxe,
  Junteng Jia, Kalyan~Vasuden Alwala, Kartikeya Upasani, Kate Plawiak, Ke~Li,
  Kenneth Heafield, Kevin Stone, et~al. 2024.
\newblock \href {https://doi.org/10.48550/ARXIV.2407.21783} {The llama 3 herd
  of models}.
\newblock \emph{CoRR}, abs/2407.21783.

\bibitem[{Durmus et~al.(2024)Durmus, Nguyen, Liao, Schiefer, Askell, Bakhtin,
  Chen, Hatfield-Dodds, Hernandez, Joseph, Lovitt, McCandlish, Sikder, Tamkin,
  Thamkul, Kaplan, Clark, and Ganguli}]{durmus2024towards}
Esin Durmus, Karina Nguyen, Thomas Liao, Nicholas Schiefer, Amanda Askell,
  Anton Bakhtin, Carol Chen, Zac Hatfield-Dodds, Danny Hernandez, Nicholas
  Joseph, Liane Lovitt, Sam McCandlish, Orowa Sikder, Alex Tamkin, Janel
  Thamkul, Jared Kaplan, Jack Clark, and Deep Ganguli. 2024.
\newblock \href {https://openreview.net/forum?id=zl16jLb91v} {Towards measuring
  the representation of subjective global opinions in language models}.
\newblock In \emph{First Conference on Language Modeling}.

\bibitem[{Eikema and Aziz(2020)}]{eikema-aziz-2020-map}
Bryan Eikema and Wilker Aziz. 2020.
\newblock \href {https://doi.org/10.18653/v1/2020.coling-main.398} {Is {MAP}
  decoding all you need? the inadequacy of the mode in neural machine
  translation}.
\newblock In \emph{Proceedings of the 28th International Conference on
  Computational Linguistics}, pages 4506--4520, Barcelona, Spain (Online).
  International Committee on Computational Linguistics.

\bibitem[{EVS(2022)}]{evs}
EVS. 2022.
\newblock \href {https://doi.org/10.4232/1.14021} {Evs trend file 1981-2017}.
\newblock GESIS, Cologne. ZA7503 Data file Version 3.0.0,
  https://doi.org/10.4232/1.14021.

\bibitem[{Feng et~al.(2023)Feng, Park, Liu, and
  Tsvetkov}]{feng-etal-2023-pretraining}
Shangbin Feng, Chan~Young Park, Yuhan Liu, and Yulia Tsvetkov. 2023.
\newblock \href {https://doi.org/10.18653/v1/2023.acl-long.656} {From
  pretraining data to language models to downstream tasks: Tracking the trails
  of political biases leading to unfair {NLP} models}.
\newblock In \emph{Proceedings of the 61st Annual Meeting of the Association
  for Computational Linguistics (Volume 1: Long Papers)}, pages 11737--11762,
  Toronto, Canada. Association for Computational Linguistics.

\bibitem[{Graham et~al.(2011)Graham, Nosek, Haidt, Iyer, Koleva, and
  Ditto}]{graham2011MFQ1}
Jesse Graham, Brian Nosek, Jonathan Haidt, Ravi Iyer, Sena~P Koleva, and
  Peter~H Ditto. 2011.
\newblock Mapping the moral domain.
\newblock \emph{Journal of Personality and Social Psychology}, 101
  (2):366–385.

\bibitem[{Gurgurov et~al.(2025)Gurgurov, Trinley, Vykopal, van Genabith,
  Ostermann, and Zamparelli}]{gurgurov2025multilingual}
Daniil Gurgurov, Katharina Trinley, Ivan Vykopal, Josef van Genabith, Simon
  Ostermann, and Roberto Zamparelli. 2025.
\newblock \href {https://doi.org/10.48550/ARXIV.2507.22623} {Multilingual
  political views of large language models: Identification and steering}.
\newblock \emph{CoRR}, abs/2507.22623.

\bibitem[{Haerpfer et~al.(2020)Haerpfer, Inglehart, Moreno, Welzel, Kizilova,
  Diez-Medrano, Lagos, Norris, Ponarin, and Puranen}]{haerpfer2020world}
Christian Haerpfer, Ronald Inglehart, Alejandro Moreno, Christian Welzel,
  Kseniya Kizilova, Jaime Diez-Medrano, Marta Lagos, Pippa Norris, Eduard
  Ponarin, and Bi~Puranen. 2020.
\newblock World values survey wave 7 (2017-2020) cross-national data-set.
\newblock \emph{(No Title)}.

\bibitem[{Hofstede(1984)}]{hofstede1984culture}
Geert Hofstede. 1984.
\newblock \href
  {https://books.google.dk/books?hl=zh-CN&lr=&id=Cayp_Um4O9gC&oi=fnd&pg=PA13&dq=Culture\%27s+consequences:+International+differences+in+work-related+values&ots=V5IFIzMIH5&sig=mhCD4hJXwu5lj239ZcXyGquJIcw&redir_esc=y#v=onepage&q=Culture's\%20consequences\%3A\%20International\%20differences\%20in\%20work-related\%20values&f=false}
  {\emph{Culture's consequences: International differences in work-related
  values}}, volume~5.
\newblock sage.

\bibitem[{Jiang et~al.(2023)Jiang, Sablayrolles, Mensch, Bamford, Chaplot,
  de~Las~Casas, Bressand, Lengyel, Lample, Saulnier, Lavaud, Lachaux, Stock,
  Scao, Lavril, Wang, Lacroix, and Sayed}]{mistral7b}
Albert~Q. Jiang, Alexandre Sablayrolles, Arthur Mensch, Chris Bamford,
  Devendra~Singh Chaplot, Diego de~Las~Casas, Florian Bressand, Gianna Lengyel,
  Guillaume Lample, Lucile Saulnier, L{\'{e}}lio~Renard Lavaud, Marie{-}Anne
  Lachaux, Pierre Stock, Teven~Le Scao, Thibaut Lavril, Thomas Wang,
  Timoth{\'{e}}e Lacroix, and William~El Sayed. 2023.
\newblock \href {https://doi.org/10.48550/ARXIV.2310.06825} {Mistral 7b}.
\newblock \emph{CoRR}, abs/2310.06825.

\bibitem[{Kazemi et~al.(2024)Kazemi, Gerhardt, Katz, Kuria, Pan, and
  Prabhakar}]{martinez2024cultural}
Sharif Kazemi, Gloria Gerhardt, Jonty Katz, Caroline~Ida Kuria, Estelle Pan,
  and Umang Prabhakar. 2024.
\newblock \href {https://doi.org/10.48550/ARXIV.2410.10489} {Cultural fidelity
  in large-language models: An evaluation of online language resources as a
  driver of model performance in value representation}.
\newblock \emph{CoRR}, abs/2410.10489.

\bibitem[{Kim and Baek(2024)}]{kim2024exploring}
Minsang Kim and Seungjun Baek. 2024.
\newblock \href {https://doi.org/10.48550/ARXIV.2412.08846} {Exploring large
  language models on cross-cultural values in connection with training
  methodology}.
\newblock \emph{CoRR}, abs/2412.08846.

\bibitem[{Kumar and Joshi(2022)}]{kumar-joshi-2022-striking}
Ashutosh Kumar and Aditya Joshi. 2022.
\newblock \href {https://doi.org/10.18653/v1/2022.findings-acl.148} {Striking a
  balance: Alleviating inconsistency in pre-trained models for symmetric
  classification tasks}.
\newblock In \emph{Findings of the Association for Computational Linguistics:
  ACL 2022}, pages 1887--1895, Dublin, Ireland. Association for Computational
  Linguistics.

\bibitem[{Lederman and Mahowald(2024)}]{lederman-mahowald-2024-language}
Harvey Lederman and Kyle Mahowald. 2024.
\newblock \href {https://doi.org/10.1162/tacl_a_00690} {Are language models
  more like libraries or like librarians? bibliotechnism, the novel reference
  problem, and the attitudes of {LLM}s}.
\newblock \emph{Transactions of the Association for Computational Linguistics},
  12:1087--1103.

\bibitem[{Liu et~al.(2024)Liu, Zhang, Yan, Wu, Yang, and Lu}]{liu2024measuring}
Songyuan Liu, Ziyang Zhang, Runze Yan, Wei Wu, Carl Yang, and Jiaying Lu. 2024.
\newblock \href {https://doi.org/10.48550/ARXIV.2410.11647} {Measuring
  spiritual values and bias of large language models}.
\newblock \emph{CoRR}, abs/2410.11647.

\bibitem[{Martins et~al.(2024)Martins, Fernandes, Alves, Guerreiro, Rei, Alves,
  Pombal, Farajian, Faysse, Klimaszewski, Colombo, Haddow, de~Souza, Birch, and
  Martins}]{martins2024eurollm}
Pedro~Henrique Martins, Patrick Fernandes, Jo{\~{a}}o Alves, Nuno~Miguel
  Guerreiro, Ricardo Rei, Duarte~M. Alves, Jos{\'{e}} Pombal, M.~Amin Farajian,
  Manuel Faysse, Mateusz Klimaszewski, Pierre Colombo, Barry Haddow, Jos{\'{e}}
  G.~C. de~Souza, Alexandra Birch, and Andr{\'{e}} F.~T. Martins. 2024.
\newblock \href {https://doi.org/10.48550/ARXIV.2409.16235} {Eurollm:
  Multilingual language models for europe}.
\newblock \emph{CoRR}, abs/2409.16235.

\bibitem[{Mather et~al.(2005)Mather, Rivers, and Jacobsen}]{mather2005american}
Mark Mather, Kerri~L Rivers, and Linda~A Jacobsen. 2005.
\newblock The american community survey.
\newblock \emph{Population Bulletin}, 60(3).

\bibitem[{Moore et~al.(2024)Moore, Deshpande, and Yang}]{moore-etal-2024-large}
Jared Moore, Tanvi Deshpande, and Diyi Yang. 2024.
\newblock \href {https://doi.org/10.18653/v1/2024.findings-emnlp.891} {Are
  large language models consistent over value-laden questions?}
\newblock In \emph{Findings of the Association for Computational Linguistics:
  EMNLP 2024}, pages 15185--15221, Miami, Florida, USA. Association for
  Computational Linguistics.

\bibitem[{Motoki et~al.(2024)Motoki, Pinho~Neto, and
  Rodrigues}]{motoki2024more}
Fabio Motoki, Valdemar Pinho~Neto, and Victor Rodrigues. 2024.
\newblock More human than human: measuring chatgpt political bias.
\newblock \emph{Public Choice}, 198(1):3--23.

\bibitem[{Nunes et~al.(2024)Nunes, Almeida, Araujo, and
  Barbosa}]{nunes2024large}
José~Luiz Nunes, Guilherme F. C.~F. Almeida, Marcelo~de Araujo, and Simone
  D.~J. Barbosa. 2024.
\newblock \href {https://doi.org/10.1609/aies.v7i1.31704} {Are large language
  models moral hypocrites? a study based on moral foundations}.
\newblock \emph{Proceedings of the AAAI/ACM Conference on AI, Ethics, and
  Society}, 7(1):1074--1087.

\bibitem[{Olmedo et~al.(2023)Olmedo, Hardt, and
  Mendler{-}D{\"{u}}nner}]{olmedo2023questioning}
Ricardo~Dominguez Olmedo, Moritz Hardt, and Celestine Mendler{-}D{\"{u}}nner.
  2023.
\newblock \href {https://doi.org/10.48550/ARXIV.2306.07951} {Questioning the
  survey responses of large language models}.
\newblock \emph{CoRR}, abs/2306.07951.

\bibitem[{Qu and Wang(2024)}]{wang2024performance}
Yao Qu and Jue Wang. 2024.
\newblock Performance and biases of large language models in public opinion
  simulation.
\newblock \emph{Humanities and Social Sciences Communications}, 11(1):1--13.

\bibitem[{R{\"o}ttger et~al.(2024)R{\"o}ttger, Hofmann, Pyatkin, Hinck, Kirk,
  Schuetze, and Hovy}]{rottger-etal-2024-political}
Paul R{\"o}ttger, Valentin Hofmann, Valentina Pyatkin, Musashi Hinck, Hannah
  Kirk, Hinrich Schuetze, and Dirk Hovy. 2024.
\newblock \href {https://doi.org/10.18653/v1/2024.acl-long.816} {Political
  compass or spinning arrow? towards more meaningful evaluations for values and
  opinions in large language models}.
\newblock In \emph{Proceedings of the 62nd Annual Meeting of the Association
  for Computational Linguistics (Volume 1: Long Papers)}, pages 15295--15311,
  Bangkok, Thailand. Association for Computational Linguistics.

\bibitem[{Rupprecht et~al.(2025)Rupprecht, Ahnert, and
  Strohmaier}]{chen2025prompt}
Jens Rupprecht, Georg Ahnert, and Markus Strohmaier. 2025.
\newblock \href {https://doi.org/10.48550/ARXIV.2507.07188} {Prompt
  perturbations reveal human-like biases in {LLM} survey responses}.
\newblock \emph{CoRR}, abs/2507.07188.

\bibitem[{Sanders et~al.(2023)Sanders, Ulinich, and
  Schneier}]{sanders2023demonstrations}
Nathan~E. Sanders, Alex Ulinich, and Bruce Schneier. 2023.
\newblock Demonstrations of the potential of {AI}-based political issue
  polling.
\newblock \emph{Harvard Data Science Review}, 5(4).
\newblock Https://hdsr.mitpress.mit.edu/pub/dm2hrtx0.

\bibitem[{Santurkar et~al.(2023)Santurkar, Durmus, Ladhak, Lee, Liang, and
  Hashimoto}]{santurkar2023opinions}
Shibani Santurkar, Esin Durmus, Faisal Ladhak, Cinoo Lee, Percy Liang, and
  Tatsunori Hashimoto. 2023.
\newblock \href {https://proceedings.mlr.press/v202/santurkar23a.html} {Whose
  opinions do language models reflect?}
\newblock In \emph{International Conference on Machine Learning, {ICML} 2023,
  23-29 July 2023, Honolulu, Hawaii, {USA}}, volume 202 of \emph{Proceedings of
  Machine Learning Research}, pages 29971--30004. {PMLR}.

\bibitem[{Scherrer et~al.(2023)Scherrer, Shi, Feder, and
  Blei}]{scherrer2023evaluating}
Nino Scherrer, Claudia Shi, Amir Feder, and David~M. Blei. 2023.
\newblock Evaluating the moral beliefs encoded in llms.
\newblock In \emph{Advances in Neural Information Processing Systems 36: Annual
  Conference on Neural Information Processing Systems 2023, NeurIPS 2023}, New
  Orleans, LA, USA.

\bibitem[{Schwartz(1992)}]{schwartz1992universals}
Shalom~H Schwartz. 1992.
\newblock Universals in the content and structure of values: Theoretical
  advances and empirical tests in 20 countries.
\newblock In \emph{Advances in experimental social psychology}, volume~25,
  pages 1--65. Elsevier.

\bibitem[{Shen et~al.(2024)Shen, Knearem, Ghosh, Yang, Mitra, and
  Huang}]{shen2024valuecompass}
Hua Shen, Tiffany Knearem, Reshmi Ghosh, Yu{-}Ju Yang, Tanushree Mitra, and Yun
  Huang. 2024.
\newblock \href {https://doi.org/10.48550/ARXIV.2409.09586} {Valuecompass: {A}
  framework of fundamental values for human-ai alignment}.
\newblock \emph{CoRR}, abs/2409.09586.

\bibitem[{Sukiennik et~al.(2025)Sukiennik, Gao, Xu, and
  Li}]{sukiennik2024evaluation}
Nicholas Sukiennik, Chen Gao, Fengli Xu, and Yong Li. 2025.
\newblock \href {https://doi.org/10.48550/ARXIV.2504.08863} {An evaluation of
  cultural value alignment in {LLM}}.
\newblock \emph{CoRR}, abs/2504.08863.

\bibitem[{Tao et~al.(2024)Tao, Viberg, Baker, and Kizilcec}]{tao2024cultural}
Yan Tao, Olga Viberg, Ryan~S Baker, and René~F Kizilcec. 2024.
\newblock \href {https://doi.org/10.1093/pnasnexus/pgae346} {Cultural bias and
  cultural alignment of large language models}.
\newblock \emph{PNAS Nexus}, 3(9):pgae346.

\bibitem[{Vida et~al.(2024)Vida, Damken, and Lauscher}]{vida2024decoding}
Karina Vida, Fabian Damken, and Anne Lauscher. 2024.
\newblock \href {https://doi.org/10.1609/aies.v7i1.31741} {Decoding
  multilingual moral preferences: Unveiling llm’s biases through the moral
  machine experiment}.
\newblock \emph{Proceedings of the AAAI/ACM Conference on AI, Ethics, and
  Society}, 7(1):1490--1501.

\bibitem[{Wang et~al.(2024)Wang, Ma, Hu, Weber-Genzel, R{\"o}ttger, Kreuter,
  Hovy, and Plank}]{wang-etal-2024-answer-c}
Xinpeng Wang, Bolei Ma, Chengzhi Hu, Leon Weber-Genzel, Paul R{\"o}ttger,
  Frauke Kreuter, Dirk Hovy, and Barbara Plank. 2024.
\newblock \href {https://doi.org/10.18653/v1/2024.findings-acl.441} {{``}my
  answer is {C}{''}: First-token probabilities do not match text answers in
  instruction-tuned language models}.
\newblock In \emph{Findings of the Association for Computational Linguistics:
  ACL 2024}, pages 7407--7416, Bangkok, Thailand. Association for Computational
  Linguistics.

\bibitem[{Wei et~al.(2022)Wei, Wang, Schuurmans, Bosma, Ichter, Xia, Chi, Le,
  and Zhou}]{cot}
Jason Wei, Xuezhi Wang, Dale Schuurmans, Maarten Bosma, Brian Ichter, Fei Xia,
  Ed~H. Chi, Quoc~V. Le, and Denny Zhou. 2022.
\newblock \href
  {http://papers.nips.cc/paper\_files/paper/2022/hash/9d5609613524ecf4f15af0f7b31abca4-Abstract-Conference.html}
  {Chain-of-thought prompting elicits reasoning in large language models}.
\newblock In \emph{Advances in Neural Information Processing Systems 35: Annual
  Conference on Neural Information Processing Systems 2022, NeurIPS 2022, New
  Orleans, LA, USA, November 28 - December 9, 2022}.

\bibitem[{Wiher et~al.(2022)Wiher, Meister, and
  Cotterell}]{wiher-etal-2022-decoding}
Gian Wiher, Clara Meister, and Ryan Cotterell. 2022.
\newblock \href {https://doi.org/10.1162/tacl_a_00502} {On decoding strategies
  for neural text generators}.
\newblock \emph{Transactions of the Association for Computational Linguistics},
  10:997--1012.

\bibitem[{Wright et~al.(2024)Wright, Arora, Borenstein, Yadav, Belongie, and
  Augenstein}]{wright-etal-2024-llm}
Dustin Wright, Arnav Arora, Nadav Borenstein, Srishti Yadav, Serge Belongie,
  and Isabelle Augenstein. 2024.
\newblock \href {https://doi.org/10.18653/v1/2024.findings-emnlp.995} {{LLM}
  tropes: Revealing fine-grained values and opinions in large language models}.
\newblock In \emph{Findings of the Association for Computational Linguistics:
  EMNLP 2024}, pages 17085--17112, Miami, Florida, USA. Association for
  Computational Linguistics.

\bibitem[{Xu et~al.(2024)Xu, Leng, Yu, and Xiong}]{xu2024self}
Shaoyang Xu, Yongqi Leng, Linhao Yu, and Deyi Xiong. 2024.
\newblock \href {https://doi.org/10.48550/ARXIV.2410.12971} {Self-pluralising
  culture alignment for large language models}.
\newblock \emph{CoRR}, abs/2410.12971.

\bibitem[{Yang et~al.(2024)Yang, Yang, Zhang, Hui, Zheng, Yu, Li, Liu, Huang,
  Wei, Lin, Yang, Tu, Zhang, Yang, Yang, Zhou, Lin, Dang, Lu, Bao, Yang, Yu,
  Li, Xue, Zhang, Zhu, Men, Lin, Li, Xia, Ren, Ren, Fan, Su, Zhang, Wan, Liu,
  Cui, Zhang, and Qiu}]{yang2024qwen}
An~Yang, Baosong Yang, Beichen Zhang, Binyuan Hui, Bo~Zheng, Bowen Yu,
  Chengyuan Li, Dayiheng Liu, Fei Huang, Haoran Wei, Huan Lin, Jian Yang,
  Jianhong Tu, Jianwei Zhang, Jianxin Yang, Jiaxi Yang, Jingren Zhou, Junyang
  Lin, Kai Dang, Keming Lu, Keqin Bao, Kexin Yang, Le~Yu, Mei Li, Mingfeng Xue,
  Pei Zhang, Qin Zhu, Rui Men, Runji Lin, Tianhao Li, Tingyu Xia, Xingzhang
  Ren, Xuancheng Ren, Yang Fan, Yang Su, Yichang Zhang, Yu~Wan, Yuqiong Liu,
  Zeyu Cui, Zhenru Zhang, and Zihan Qiu. 2024.
\newblock \href {https://doi.org/10.48550/ARXIV.2412.15115} {Qwen2.5 technical
  report}.
\newblock \emph{CoRR}, abs/2412.15115.

\end{thebibliography}

\appendix

\section{Example of Prompt Formulation}
\label{sec:appendix}

\subsection*{Question 1: Direct Prompt}

\emph{English}: For each of the following, indicate how important it is in human life. 1 is very important, 4 is not important at all. Say a number from 1 to 4 and nothing else. Family.

\emph{German}: Geben Sie für jeden der folgenden Punkte an, wie wichtig er im menschlichen Leben ist. 1 ist sehr wichtig, 4 ist überhaupt nicht wichtig. Nennen Sie die Zahl von 1 bis 4 und sonst nichts. Familie.

\emph{Czech}: U každé z následujících označte, jak je důležitá v lidském životě. 1 je velmi důležitá, 4 není vůbec důležitá. Napiš číslo od 1 do 4 a nic jiného. Rodina.

\subsection*{Question 1: CoT Prompt}

\emph{English}:  For each of the following, indicate how important it is in human life. 1 is very important, 4 is not important at all. First, provide a step-by-step justification, and finally, say the number from 1 to 4 and nothing else. Family.

\emph{German}: Geben Sie für jeden der folgenden Punkte an, wie wichtig er im menschlichen Leben ist. 1 ist sehr wichtig, 4 ist überhaupt nicht wichtig. Geben Sie zunächst eine schrittweise Begründung an und nennen Sie abschließend die Zahl von 1 bis 4 und sonst nichts. Familie.

\emph{Czech}: U každé z následujících označte, jak je důležitá v lidském životě. 1 je velmi důležitá, 4 není vůbec důležitá. Nejprve napiš zdůvodnění krok za krokem a nakonec napiš číslo od 1 do 4 a nic jiného. Rodina.

\subsection*{Question 41: Direct Prompt}

\emph{English}: For the following statement, tell me how much agreeable it is. 1 is strongly agree 3 is neutral, 5 is strongly disagree. Say a number from 1 to 5 and nothing else. Work should always come first, even if it means less spare time.

\emph{German}: Sagen Sie mir, wie sehr kann man der folgenden Aussage zustimmen. 1 bedeutet völlige Zustimmung, 3 bedeutet neutral, 5 bedeutet überhaupt keine Zustimmung. Nennen Sie eine Zahl von 1 und 5 und sonst nichts. Die Arbeit sollte immer an erster Stelle stehen, auch wenn dies weniger Freizeit bedeutet.

\emph{Czech}: U následující tvrzení uveď, jak moc s ním lze souhlasit. 1 úplný souhlas, 3 je neutrální, 5 naprostý nesouhlas. Napiš číslo od 1 do 5 a nic jiného. Práce by měla být vždy na prvním místě, i když to znamená méně volného času.

\subsection*{Question 41: CoT Prompt}

\emph{English}: For the following statement, tell me how much agreeable it is. 1 is strongly agree 3 is neutral, 5 is strongly disagree. First provide a step-by-step justification and finally say a number from 1 to 5 and nothing else. Work should always come first, even if it means less spare time.

\emph{German}: Sagen Sie mir, wie sehr kann man der folgenden Aussage zustimmen. 1 bedeutet völlige Zustimmung, 3 bedeutet neutral, 5 bedeutet überhaupt keine Zustimmung. Geben Sie zunächst eine schrittweise Begründung an und nennen Sie abschließend eine Zahl von 1 und 5 und sonst nichts. Die Arbeit sollte immer an erster Stelle stehen, auch wenn dies weniger Freizeit bedeutet.

\emph{Czech}: U následující tvrzení uveď, jak moc s ním lze souhlasit. 1 úplný souhlas, 3 je neutrální, 5 naprostý nesouhlas. Nejprve napiš zdůvodnění krok za krokem a nakonec napiš číslo od 1 do 5 a nic jiného. Práce by měla být vždy na prvním místě, i když to znamená méně volného času.

\section{Detailed Multilingual Results}

Here, we present the MSD, KL Divergence, Correlation norm, and Self-correlation distance for all combinations of languages (English: en, German: de, Czech: cs) and all countries (United States: USA, United Kingdom: GBR, Czechia: CZE, Germany: DEU, Iran: IRN, China: CHN) in Table~\ref{tab:country_compare}. Table~\ref{tab:usa_compare} is a subset of this table. Table~\ref{tab:match_countires} with point-biserial correlation of country-language matching and evaluation metrics and Table~\ref{tab:metric_correlation} are computed from numbers in Table~\ref{tab:country_compare}. Table~\ref{tab:metric_correlation_breakdown} shows the metric correlation separately from LLaMA~3 and Mistral~2.

Table~\ref{tab:country_compare} shows country comparison using the data from WVS with the metrics that we use in the paper. In the paper, we occasionally compare the model alignment to differences between countries. For this, we use data from this table.

\begin{table*}[t]
\setlength\tabcolsep{2.0pt}
\footnotesize\centering

\begin{minipage}{.25\textwidth}
\centering
\begin{tabular}{l ccc}
\toprule
 & MSD & KLD & CorrD \\ \midrule
MSD & --- & \hphantom{-}.276 & -.368 \\
KLD & \hphantom{-}.389 & ---  & -.064 \\
CorrD & -.342 & -.005 & --- \\
\bottomrule
\end{tabular} \\[.5ex]
(a) LLaMA~3
\end{minipage}%
\begin{minipage}{.25\textwidth}
\centering
\begin{tabular}{l ccc}
\toprule
 & MSD & KLD & CorrD \\ \midrule
MSD & --- & \hphantom{-}.832 & -.657 \\
KLD & \hphantom{-}.925 & --- & -.500 \\
CorrD & -.666 & -.526 & --- \\
\bottomrule
\end{tabular} \\[.5ex]
(b) Mistral~2
\end{minipage}%
\begin{minipage}{.25\textwidth}
\centering
\begin{tabular}{lccc}
\toprule
 & MSD & KLD & CorrD \\
\midrule
MSD & --- &.691 &.279 \\
KLD &.724 & --- &.251 \\
CorrD &.328 &.271 & --- \\
\bottomrule
\end{tabular} \\[.5ex]
(c) EuroLLM
\end{minipage}%
\begin{minipage}{.25\textwidth}
\centering
\begin{tabular}{lccc}
\toprule
 & MSD & KLD & CorrD \\
\midrule
MSD & --- &.617 &.120 \\
KLD &.656 & --- &.495 \\
CorrD &.009 &.396 & --- \\
\bottomrule
\end{tabular} \\[.5ex]
(d) Qwen 2.5
\end{minipage}%

\caption{Breakdown of correlation of the metrics (Pearson above the diagonal, Spearman under the diagonal) over all languages and countries for (a) LLaMA~3, (b) Mistral~2, (c) EuroLLM, and (d) Qwen~2.5}\label{tab:metric_correlation_breakdown}

\end{table*}

\begin{table*}[ht]

\scriptsize\centering
\setlength\tabcolsep{3.0pt}

\begin{tabular}{l@{\hskip 5pt} lll cccccc cccccc c cccccc}
\toprule
\multirow{2}{*}{\rotatebox{90}{Model}} & \multirow{2}{*}{\begin{minipage}{.8cm}Prompt type\end{minipage}} & \multirow{2}{*}{\begin{minipage}{.2cm}De\-co\-de\end{minipage}} & \multirow{2}{*}{\rotatebox{90}{Lng~~}} & \multicolumn{6}{c}{Mean Sq. Difference} & \multicolumn{6}{c}{KL Divergence} & \multirow{2}{*}{\begin{minipage}{.6cm}\centering Corr. norm\end{minipage}} & \multicolumn{6}{c}{Self-correlation distance} \\ \cmidrule(lr){5-10} \cmidrule(lr){11-16} \cmidrule(lr){18-23}
& & & & USA & GBR & CZE & DEU & IRN & CHN    & USA & GBR & CZE & DEU & IRN & CHN   & & USA & GBR & CZE & DEU & IRN & CHN
\\ \midrule

\multirow{12}{*}{\rotatebox{90}{LLaMA~3\quad\quad}}
& \multirow{6}{*}{\begin{minipage}{.8cm}Score only\end{minipage}} & \multirow{3}{*}{Gr.} & en &
\mseCell{.098} & \mseCell{.098} & \mseCell{.110} & \mseCell{.096} & \mseCell{.115} & \mseCell{.102} & \klCell{1.71} & \klCell{1.66} & \klCell{1.74} & \klCell{1.67} & \klCell{1.81} & \klCell{1.70}  \\
& & & de &
\mseCell{.084} & \mseCell{.076} & \mseCell{.103} & \mseCell{.079} & \mseCell{.132} & \mseCell{.112} & \klCell{1.65} & \klCell{1.56} & \klCell{1.70} & \klCell{1.62} & \klCell{1.93} & \klCell{1.85} \\
& & & cs &
\mseCell{.086} & \mseCell{.071} & \mseCell{.110} & \mseCell{.077} & \mseCell{.145} & \mseCell{.119} & \klCell{1.56} & \klCell{1.41} & \klCell{1.64} & \klCell{1.41} & \klCell{1.83} & \klCell{1.69} \\ \cmidrule{3-23}
& & \multirow{3}{*}{Spl.} & en &
\mseCell{.073} & \mseCell{.073} & \mseCell{.083} & \mseCell{.073} & \mseCell{.090} & \mseCell{.083} & \klCell{2.99} & \klCell{2.90} & \klCell{3.26} & \klCell{2.81} & \klCell{3.43} & \klCell{3.23} & \normCell{0.90} & \corrCell{1.26} & \corrCell{1.20} & \corrCell{1.17} & \corrCell{0.95} & \corrCell{0.96} & \corrCell{0.93} \\
& & & de &
\mseCell{.084} & \mseCell{.073} & \mseCell{.099} & \mseCell{.079} & \mseCell{.127} & \mseCell{.114} & \klCell{3.58} & \klCell{3.33} & \klCell{3.76} & \klCell{3.30} & \klCell{4.14} & \klCell{3.79} & \normCell{0.67} & \corrCell{1.31} & \corrCell{1.25} & \corrCell{1.24} & \corrCell{1.01} & \corrCell{0.99} & \corrCell{1.01} \\
& & & cs &
\mseCell{.067} & \mseCell{.047} & \mseCell{.080} & \mseCell{.057} & \mseCell{.114} & \mseCell{.105} & \klCell{2.90} & \klCell{2.49} & \klCell{3.05} & \klCell{2.46} & \klCell{3.52} & \klCell{3.04} & \normCell{0.78} & \corrCell{1.26} & \corrCell{1.20} & \corrCell{1.20} & \corrCell{0.96} & \corrCell{0.94} & \corrCell{0.95} \\ \cmidrule{2-23}
& \multirow{6}{*}{CoT} & \multirow{3}{*}{Gr.} & en &
\mseCell{.088} & \mseCell{.083} & \mseCell{.112} & \mseCell{.090} & \mseCell{.152} & \mseCell{.133} & \klCell{1.68} & \klCell{1.57} & \klCell{1.73} & \klCell{1.63} & \klCell{2.01} & \klCell{1.89} \\
& & & de &
\mseCell{.085} & \mseCell{.072} & \mseCell{.104} & \mseCell{.080} & \mseCell{.134} & \mseCell{.117} & \klCell{1.61} & \klCell{1.47} & \klCell{1.69} & \klCell{1.55} & \klCell{1.96} & \klCell{1.82} \\
& & & cs &
\mseCell{.071} & \mseCell{.060} & \mseCell{.089} & \mseCell{.068} & \mseCell{.125} & \mseCell{.104} & \klCell{1.57} & \klCell{1.46} & \klCell{1.63} & \klCell{1.58} & \klCell{1.87} & \klCell{1.71} \\ \cmidrule{3-23}
& & \multirow{3}{*}{Spl.} & en &
\mseCell{.059} & \mseCell{.046} & \mseCell{.075} & \mseCell{.054} & \mseCell{.118} & \mseCell{.111} & \klCell{1.47} & \klCell{1.30} & \klCell{1.70} & \klCell{1.29} & \klCell{1.94} & \klCell{1.93} & \normCell{1.70} & \corrCell{1.29} & \corrCell{1.26} & \corrCell{1.22} & \corrCell{1.14} & \corrCell{1.19} & \corrCell{1.03} \\
& & & de &
\mseCell{.054} & \mseCell{.042} & \mseCell{.063} & \mseCell{.050} & \mseCell{.105} & \mseCell{.094} & \klCell{1.05} & \klCell{0.87} & \klCell{1.12} & \klCell{0.87} & \klCell{1.35} & \klCell{1.34} & \normCell{1.50} & \corrCell{1.17} & \corrCell{1.14} & \corrCell{1.10} & \corrCell{0.97} & \corrCell{1.02} & \corrCell{0.89} \\
& & & cs &
\mseCell{.049} & \mseCell{.035} & \mseCell{.059} & \mseCell{.042} & \mseCell{.095} & \mseCell{.087} & \klCell{0.98} & \klCell{0.81} & \klCell{1.11} & \klCell{0.84} & \klCell{1.37} & \klCell{1.27} & \normCell{1.60} & \corrCell{1.24} & \corrCell{1.19} & \corrCell{1.14} & \corrCell{1.04} & \corrCell{1.11} & \corrCell{0.93} \\
\midrule
\multirow{12}{*}{\rotatebox{90}{Mistral2\quad\quad}}
& \multirow{6}{*}{\begin{minipage}{.8cm}Score only\end{minipage}} & \multirow{3}{*}{Gr.} & en &
\mseCell{.094} & \mseCell{.095} & \mseCell{.114} & \mseCell{.110} & \mseCell{.121} & \mseCell{.114} & \klCell{1.80} & \klCell{1.75} & \klCell{1.76} & \klCell{1.90} & \klCell{2.01} & \klCell{1.88}  \\
& & & de &
\mseCell{.096} & \mseCell{.107} & \mseCell{.105} & \mseCell{.110} & \mseCell{.100} & \mseCell{.105} & \klCell{1.90} & \klCell{1.97} & \klCell{1.82} & \klCell{2.08} & \klCell{2.08} & \klCell{1.94} \\
& & & cs &
\mseCell{.114} & \mseCell{.129} & \mseCell{.121} & \mseCell{.131} & \mseCell{.111} & \mseCell{.118} & \klCell{1.91} & \klCell{1.94} & \klCell{1.83} & \klCell{2.06} & \klCell{2.07} & \klCell{1.98} \\ \cmidrule{3-23}
& & \multirow{3}{*}{Spl.} & en &
\mseCell{.041} & \mseCell{.040} & \mseCell{.051} & \mseCell{.051} & \mseCell{.072} & \mseCell{.056} & \klCell{0.77} & \klCell{0.70} & \klCell{0.82} & \klCell{0.82} & \klCell{1.41} & \klCell{0.97} & \normCell{1.77} & \corrCell{1.17} & \corrCell{1.18} & \corrCell{1.10} & \corrCell{1.06} & \corrCell{1.10} & \corrCell{0.95} \\
& & & de &
\mseCell{.035} & \mseCell{.039} & \mseCell{.040} & \mseCell{.051} & \mseCell{.072} & \mseCell{.068} & \klCell{0.56} & \klCell{0.57} & \klCell{0.51} & \klCell{0.73} & \klCell{0.94} & \klCell{0.79} & \normCell{2.26} & \corrCell{1.37} & \corrCell{1.35} & \corrCell{1.32} & \corrCell{1.32} & \corrCell{1.37} & \corrCell{1.19} \\
& & & cs &
\mseCell{.034} & \mseCell{.036} & \mseCell{.037} & \mseCell{.046} & \mseCell{.060} & \mseCell{.058} & \klCell{0.41} & \klCell{0.41} & \klCell{0.40} & \klCell{0.49} & \klCell{0.62} & \klCell{0.55} & \normCell{2.64} & \corrCell{1.53} & \corrCell{1.44} & \corrCell{1.51} & \corrCell{1.58} & \corrCell{1.67} & \corrCell{1.46} \\ \cmidrule{2-23}
& \multirow{6}{*}{CoT} & \multirow{3}{*}{Gr.} & en &
\mseCell{.188} & \mseCell{.181} & \mseCell{.187} & \mseCell{.187} & \mseCell{.156} & \mseCell{.152} & \klCell{1.95} & \klCell{1.85} & \klCell{1.98} & \klCell{2.01} & \klCell{1.78} & \klCell{1.94} \\
& & & de &
\mseCell{.196} & \mseCell{.195} & \mseCell{.202} & \mseCell{.202} & \mseCell{.281} & \mseCell{.288} & \klCell{1.88} & \klCell{1.87} & \klCell{1.86} & \klCell{1.98} & \klCell{2.03} & \klCell{2.19} \\
& & & cs &
\mseCell{.163} & \mseCell{.156} & \mseCell{.147} & \mseCell{.169} & \mseCell{.196} & \mseCell{.189} & \klCell{1.82} & \klCell{1.80} & \klCell{1.72} & \klCell{1.88} & \klCell{1.96} & \klCell{1.88} \\ \cmidrule{3-23}
& & \multirow{3}{*}{Spl.} & en &
\mseCell{.022} & \mseCell{.025} & \mseCell{.030} & \mseCell{.035} & \mseCell{.050} & \mseCell{.048} & \klCell{0.26} & \klCell{0.26} & \klCell{0.29} & \klCell{0.33} & \klCell{0.45} & \klCell{0.43} & \normCell{2.80} & \corrCell{1.62} & \corrCell{1.56} & \corrCell{1.64} & \corrCell{1.68} & \corrCell{1.78} & \corrCell{1.58} \\
& & & de &
\mseCell{.066} & \mseCell{.082} & \mseCell{.064} & \mseCell{.088} & \mseCell{.054} & \mseCell{.062} & \klCell{0.81} & \klCell{0.77} & \klCell{0.82} & \klCell{0.77} & \klCell{0.62} & \klCell{0.81} & \normCell{3.20} & \corrCell{1.88} & \corrCell{1.76} & \corrCell{1.85} & \corrCell{1.96} & \corrCell{2.03} & \corrCell{1.80} \\
& & & cs &
\mseCell{.039} & \mseCell{.045} & \mseCell{.036} & \mseCell{.056} & \mseCell{.052} & \mseCell{.050} & \klCell{0.47} & \klCell{0.45} & \klCell{0.46} & \klCell{0.46} & \klCell{0.41} & \klCell{0.49} & \normCell{3.32} & \corrCell{1.92} & \corrCell{1.82} & \corrCell{1.92} & \corrCell{2.03} & \corrCell{2.09} & \corrCell{1.91} \\
\midrule
\multirow{12}{*}{\rotatebox{90}{EuroLLM\quad\quad}}
& \multirow{6}{*}{\begin{minipage}{.8cm}Score only\end{minipage}} & \multirow{3}{*}{Gr.} & en &
\mseCell{.165} & \mseCell{.167} & \mseCell{.193} & \mseCell{.186} & \mseCell{.162} & \mseCell{.156} & \klCell{1.91} & \klCell{1.80} & \klCell{1.97} & \klCell{1.81} & \klCell{1.72} & \klCell{1.75}  \\
& & & de &
\mseCell{.284} & \mseCell{.290} & \mseCell{.261} & \mseCell{.300} & \mseCell{.254} & \mseCell{.274} & \klCell{2.40} & \klCell{2.35} & \klCell{2.33} & \klCell{2.40} & \klCell{2.08} & \klCell{2.38} \\
& & & cs &
\mseCell{.223} & \mseCell{.244} & \mseCell{.243} & \mseCell{.248} & \mseCell{.176} & \mseCell{.209} & \klCell{2.19} & \klCell{2.21} & \klCell{2.26} & \klCell{2.22} & \klCell{1.84} & \klCell{2.14} \\ \cmidrule{3-23}
& & \multirow{3}{*}{Spl.} & en &
\mseCell{.059} & \mseCell{.069} & \mseCell{.068} & \mseCell{.081} & \mseCell{.045} & \mseCell{.055} & \klCell{0.72} & \klCell{0.65} & \klCell{0.78} & \klCell{0.68} & \klCell{0.55} & \klCell{0.75} & \normCell{2.55} & \corrCell{1.56} & \corrCell{1.54} & \corrCell{1.54} & \corrCell{1.57} & \corrCell{1.62} & \corrCell{1.44} \\
& & & de &
\mseCell{.094} & \mseCell{.111} & \mseCell{.093} & \mseCell{.121} & \mseCell{.065} & \mseCell{.080} & \klCell{1.00} & \klCell{0.95} & \klCell{1.04} & \klCell{0.93} & \klCell{0.69} & \klCell{0.92} & \normCell{2.76} & \corrCell{1.66} & \corrCell{1.59} & \corrCell{1.70} & \corrCell{1.69} & \corrCell{1.77} & \corrCell{1.57} \\
& & & cs &
\mseCell{.129} & \mseCell{.149} & \mseCell{.129} & \mseCell{.157} & \mseCell{.082} & \mseCell{.103} & \klCell{1.52} & \klCell{1.50} & \klCell{1.59} & \klCell{1.53} & \klCell{1.17} & \klCell{1.55} & \normCell{2.95} & \corrCell{1.71} & \corrCell{1.66} & \corrCell{1.78} & \corrCell{1.79} & \corrCell{1.87} & \corrCell{1.69} \\ \cmidrule{2-23}
& \multirow{6}{*}{CoT} & \multirow{3}{*}{Gr.} & en &
\mseCell{.130} & \mseCell{.130} & \mseCell{.157} & \mseCell{.142} & \mseCell{.155} & \mseCell{.170} & \klCell{1.76} & \klCell{1.69} & \klCell{1.78} & \klCell{1.76} & \klCell{1.70} & \klCell{1.82} \\
& & & de &
\mseCell{.369} & \mseCell{.402} & \mseCell{.322} & \mseCell{.393} & \mseCell{.207} & \mseCell{.279} & \klCell{0.91} & \klCell{0.97} & \klCell{0.80} & \klCell{1.08} & \klCell{0.79} & \klCell{0.94} \\
& & & cs &
\mseCell{.358} & \mseCell{.421} & \mseCell{.380} & \mseCell{.442} & \mseCell{.286} & \mseCell{.296} & \klCell{1.15} & \klCell{1.25} & \klCell{1.16} & \klCell{1.28} & \klCell{0.98} & \klCell{1.12} \\ \cmidrule{3-23}
& & \multirow{3}{*}{Spl.} & en &
\mseCell{.125} & \mseCell{.143} & \mseCell{.126} & \mseCell{.153} & \mseCell{.091} & \mseCell{.099} & \klCell{0.97} & \klCell{0.93} & \klCell{1.00} & \klCell{0.94} & \klCell{0.68} & \klCell{0.87} & \normCell{2.97} & \corrCell{1.74} & \corrCell{1.64} & \corrCell{1.78} & \corrCell{1.79} & \corrCell{1.88} & \corrCell{1.68} \\
& & & de &
\mseCell{.149} & \mseCell{.171} & \mseCell{.151} & \mseCell{.179} & \mseCell{.106} & \mseCell{.121} & \klCell{1.29} & \klCell{1.27} & \klCell{1.38} & \klCell{1.21} & \klCell{0.89} & \klCell{1.12} & \normCell{2.83} & \corrCell{1.71} & \corrCell{1.59} & \corrCell{1.74} & \corrCell{1.73} & \corrCell{1.85} & \corrCell{1.64} \\
& & & cs &
\mseCell{.120} & \mseCell{.140} & \mseCell{.119} & \mseCell{.149} & \mseCell{.086} & \mseCell{.097} & \klCell{1.12} & \klCell{1.09} & \klCell{1.14} & \klCell{1.07} & \klCell{0.76} & \klCell{0.98} & \normCell{2.84} & \corrCell{1.67} & \corrCell{1.52} & \corrCell{1.67} & \corrCell{1.71} & \corrCell{1.82} & \corrCell{1.57} \\
\midrule
\multirow{12}{*}{\rotatebox{90}{Qwen\quad\quad}}
& \multirow{6}{*}{\begin{minipage}{.8cm}Score only\end{minipage}} & \multirow{3}{*}{Gr.} & en &
\mseCell{.082} & \mseCell{.068} & \mseCell{.105} & \mseCell{.079} & \mseCell{.136} & \mseCell{.107} & \klCell{1.66} & \klCell{1.60} & \klCell{1.82} & \klCell{1.68} & \klCell{2.02} & \klCell{1.88}  \\
& & & de &
\mseCell{.070} & \mseCell{.102} & \mseCell{.069} & \mseCell{.088} & \mseCell{.092} & \mseCell{.100} & \klCell{0.82} & \klCell{0.88} & \klCell{0.73} & \klCell{0.92} & \klCell{0.92} & \klCell{0.98} \\
& & & cs &
\mseCell{.081} & \mseCell{.053} & \mseCell{.099} & \mseCell{.057} & \mseCell{.178} & \mseCell{.147} & \klCell{0.66} & \klCell{0.65} & \klCell{0.60} & \klCell{0.78} & \klCell{0.83} & \klCell{0.87} \\ \cmidrule{3-23}
& & \multirow{3}{*}{Spl.} & en &
\mseCell{.074} & \mseCell{.075} & \mseCell{.080} & \mseCell{.081} & \mseCell{.076} & \mseCell{.065} & \klCell{1.27} & \klCell{1.17} & \klCell{1.48} & \klCell{1.15} & \klCell{1.35} & \klCell{1.33} & \normCell{3.33} & \corrCell{2.13} & \corrCell{2.07} & \corrCell{2.12} & \corrCell{2.15} & \corrCell{2.16} & \corrCell{2.09} \\
& & & de &
\mseCell{.054} & \mseCell{.043} & \mseCell{.062} & \mseCell{.046} & \mseCell{.093} & \mseCell{.073} & \klCell{1.13} & \klCell{0.95} & \klCell{1.24} & \klCell{0.95} & \klCell{1.40} & \klCell{1.31} & \normCell{2.12} & \corrCell{1.61} & \corrCell{1.56} & \corrCell{1.51} & \corrCell{1.45} & \corrCell{1.50} & \corrCell{1.36} \\
& & & cs &
\mseCell{.041} & \mseCell{.033} & \mseCell{.053} & \mseCell{.040} & \mseCell{.083} & \mseCell{.060} & \klCell{0.81} & \klCell{0.69} & \klCell{0.91} & \klCell{0.65} & \klCell{1.00} & \klCell{0.91} & \normCell{1.93} & \corrCell{1.39} & \corrCell{1.33} & \corrCell{1.24} & \corrCell{1.24} & \corrCell{1.27} & \corrCell{1.16} \\ \cmidrule{2-23}
& \multirow{6}{*}{CoT} & \multirow{3}{*}{Gr.} & en &
\mseCell{.199} & \mseCell{.170} & \mseCell{.206} & \mseCell{.193} & \mseCell{.268} & \mseCell{.253} & \klCell{1.60} & \klCell{1.50} & \klCell{1.56} & \klCell{1.54} & \klCell{1.75} & \klCell{1.74} \\
& & & de &
\mseCell{.141} & \mseCell{.133} & \mseCell{.149} & \mseCell{.154} & \mseCell{.179} & \mseCell{.177} & \klCell{1.70} & \klCell{1.66} & \klCell{1.73} & \klCell{1.78} & \klCell{1.84} & \klCell{1.87} \\
& & & cs &
\mseCell{.116} & \mseCell{.113} & \mseCell{.109} & \mseCell{.101} & \mseCell{.142} & \mseCell{.153} & \klCell{1.22} & \klCell{1.21} & \klCell{1.12} & \klCell{1.25} & \klCell{1.20} & \klCell{1.35} \\ \cmidrule{3-23}
& & \multirow{3}{*}{Spl.} & en &
\mseCell{.062} & \mseCell{.048} & \mseCell{.078} & \mseCell{.058} & \mseCell{.105} & \mseCell{.078} & \klCell{1.13} & \klCell{0.97} & \klCell{1.37} & \klCell{1.01} & \klCell{1.53} & \klCell{1.39} & \normCell{1.52} & \corrCell{1.25} & \corrCell{1.13} & \corrCell{1.15} & \corrCell{1.03} & \corrCell{1.09} & \corrCell{0.94} \\
& & & de &
\mseCell{.055} & \mseCell{.050} & \mseCell{.073} & \mseCell{.061} & \mseCell{.103} & \mseCell{.093} & \klCell{0.69} & \klCell{0.59} & \klCell{0.75} & \klCell{0.65} & \klCell{0.92} & \klCell{0.84} & \normCell{1.84} & \corrCell{1.24} & \corrCell{1.16} & \corrCell{1.12} & \corrCell{1.12} & \corrCell{1.21} & \corrCell{0.98} \\
& & & cs &
\mseCell{.048} & \mseCell{.036} & \mseCell{.062} & \mseCell{.045} & \mseCell{.088} & \mseCell{.070} & \klCell{0.73} & \klCell{0.61} & \klCell{0.83} & \klCell{0.63} & \klCell{1.01} & \klCell{0.90} & \normCell{1.72} & \corrCell{1.17} & \corrCell{1.12} & \corrCell{1.10} & \corrCell{1.03} & \corrCell{1.12} & \corrCell{0.92} \\

\bottomrule
\end{tabular}

\caption{A comparison of model outputs with different prompting strategies (Score-only, CoT: Chain of Thought), decoding method (Gr.: greedy, Spl.: sampling) and different languages}\label{tab:model_compare}

\end{table*}

\begin{table*}[ht]
\scriptsize\centering
\setlength\tabcolsep{3.0pt}

\begin{tabular}{l cccccc}
\toprule
& USA & GBR & CZE & DEU & IRN & CHN \\ \midrule
USA & --- & \mseCell{.009} & \mseCell{.017} & \mseCell{.016} & \mseCell{.046} & \mseCell{.043} \\
GBR & \mseCell{.009} & --- & \mseCell{.022} & \mseCell{.011} & \mseCell{.066} & \mseCell{.052} \\
CZE & \mseCell{.017} & \mseCell{.022} & --- & \mseCell{.024} & \mseCell{.044} & \mseCell{.042} \\
DEU & \mseCell{.016} & \mseCell{.011} & \mseCell{.024} & --- & \mseCell{.069} & \mseCell{.049} \\
IRN & \mseCell{.046} & \mseCell{.066} & \mseCell{.044} & \mseCell{.069} & --- & \mseCell{.030} \\
CHN & \mseCell{.043} & \mseCell{.052} & \mseCell{.042} & \mseCell{.049} & \mseCell{.030} & --- \\
\bottomrule
\end{tabular}

\begin{tabular}{l cccccc}
\toprule
& USA & GBR & CZE & DEU & IRN & CHN \\ \midrule
USA & --- & \klCell{0.07} & \klCell{0.11} & \klCell{0.16} & \klCell{0.34} & \klCell{0.32} \\
GBR & \klCell{0.15} & --- & \klCell{0.13} & \klCell{0.09} & \klCell{0.39} & \klCell{0.35} \\
CZE & \klCell{0.17} & \klCell{0.15} & --- & \klCell{0.22} & \klCell{0.34} & \klCell{0.29} \\
DEU & \klCell{0.21} & \klCell{0.08} & \klCell{0.17} & --- & \klCell{0.36} & \klCell{0.32} \\
IRN & \klCell{0.41} & \klCell{0.42} & \klCell{0.33} & \klCell{0.44} & --- & \klCell{0.27} \\
CHN & \klCell{0.32} & \klCell{0.34} & \klCell{0.27} & \klCell{0.33} & \klCell{0.24} & --- \\
\bottomrule
\end{tabular}

\begin{tabular}{l c cccccc}
\toprule
& Norm. & USA & GBR & CZE & DEU & IRN & CHN \\ \midrule
USA & \normCell{1.66} & --- & \corrCell{0.79} & \corrCell{0.92} & \corrCell{0.78} & \corrCell{0.95} & \corrCell{0.92} \\
GBR & \normCell{1.54} & \corrCell{0.79} & --- & \corrCell{0.90} & \corrCell{0.72} & \corrCell{0.92} & \corrCell{0.78} \\
CZE & \normCell{1.55} & \corrCell{0.92} & \corrCell{0.90} & --- & \corrCell{0.83} & \corrCell{0.93} & \corrCell{0.79} \\
DEU & \normCell{1.11} & \corrCell{0.78} & \corrCell{0.72} & \corrCell{0.83} & --- & \corrCell{0.66} & \corrCell{0.64} \\
IRN & \normCell{1.06} & \corrCell{0.95} & \corrCell{0.92} & \corrCell{0.93} & \corrCell{0.66} & --- & \corrCell{0.69} \\
CHN & \normCell{1.13} & \corrCell{0.92} & \corrCell{0.78} & \corrCell{0.79} & \corrCell{0.64} & \corrCell{0.69} & --- \\
\bottomrule
\end{tabular}

\caption{Comparison of difference between countries in the World Value Survey when measured using \colorbox{CornflowerBlue!60}{mean squared difference}, \colorbox{LimeGreen!60}{KL Divergence}, the \colorbox{Goldenrod!60}{norm of the self-correlation tables} for each country and \colorbox{RedOrange!60}{distances of the self-correlation tables} across countries.}\label{tab:country_compare}

\end{table*}

\end{document}